\documentclass[journal]{IEEEtran}
\usepackage{amsmath,amsfonts}
\usepackage{algorithmic}
\usepackage{algorithm}
\usepackage{array}
\usepackage[caption=false,font=normalsize,labelfont=sf,textfont=sf]{subfig}
\usepackage{textcomp}
\usepackage{stfloats}
\usepackage{url}
\usepackage{verbatim}
\usepackage{graphicx}
\usepackage{cite}
\usepackage{xcolor}
\usepackage{booktabs}
\usepackage{multirow}
\usepackage{pifont}

\usepackage{times}  
\usepackage{helvet}  
\usepackage{courier}  

\usepackage{mathrsfs}
\usepackage{diagbox}

\begin{document}

\title{Image Captioning via Compact Bidirectional Architecture}

\author{
Zijie~Song$^*$,
Yuanen~Zhou$^{*\dagger}$,
Zhenzhen~Hu$^\dagger$,~\IEEEmembership{Member,~IEEE}, 
Daqing~Liu, Huixia~Ben,
Richang~Hong,~\IEEEmembership{Member,~IEEE},
Meng~Wang,~\IEEEmembership{Fellow,~IEEE}

\thanks{$*$ denoted equal contribution. 
$\dagger$ denoted the corresponding authors.

The work was supported by the Joint Funds of the National Natural Science Foundation of China under Grant No.~U23B2031, and the Fundamental Research Funds for the Central Universities under Grant No.~PA2025IISL0110.

Zijie Song is with the School of Big Data and Statistics, Anhui University, Hefei, 230601, China (e-mail: zjsong@ahu.edu.cn).

Yuanen Zhou is with the Institute of Artificial Intelligence, Hefei Comprehensive National Science Center, and Intelligent Interconnected Systems Laboratory of Anhui Province (Hefei University of Technology), Hefei, 230088 and 230601, China (e-mail: y.e.zhou.hb@gmail.com).

Zhenzhen Hu, Richang Hong and Meng Wang are with the School of Computer Science and Information Engineering, Hefei University of Technology, Hefei, 230601, China, (e-mail: zzhu@hfut.edu.cn; hongrc.hfut@gmail.com; eric.mengwang@gmail.com).

Daqing Liu is with JD Explore Academy, JD.com Inc, Beijing, 100176, China, (e-mail: liudq.ustc@gmail.com).

Huixia Ben is with the State Key Laboratory of Digital Intelligent Technology for Unmanned Coal Mining, Anhui University of Science and Technology, Hefei, 231131, China, (e-mail: benhuixia@aust.edu.cn).}
}


\markboth{This article has been accepted for IEEE Transactions on Multimedia.}{}

\maketitle

\begin{abstract}
 Most current image captioning models typically generate captions  from left-to-right. This unidirectional property makes them can only leverage  past context but not future context. Though refinement-based models can exploit both past and future context by generating a new caption in the  second stage based on  pre-retrieved or pre-generated captions in the first stage, the decoder of these models generally consists of two networks~(i.e. a retriever or captioner in the first stage and a captioner in the second stage), which can only be executed sequentially. In this paper, we introduce a Compact Bidirectional Transformer model for image captioning that can  leverage bidirectional context implicitly and explicitly while the decoder can be executed parallelly. Specifically, it is implemented by tightly coupling left-to-right(L2R) and right-to-left(R2L) flows into a single compact model to serve as a regularization for implicitly exploiting bidirectional context and optionally allowing explicit interaction of the bidirectional flows, while the final caption is chosen from  either L2R  or R2L flow in a sentence-level ensemble manner. We conduct extensive ablation studies on MSCOCO benchmark and find that  the compact bidirectional architecture and the sentence-level ensemble play more important roles than the explicit interaction mechanism. By combining with word-level ensemble seamlessly, the effect of sentence-level ensemble is further enlarged. We further extend the conventional one-flow self-critical training to the two-flows version under this architecture and achieve new state-of-the-art results  in comparison with non-vision-language-pretraining models. Finally, we verify the generality of this compact bidirectional architecture by extending it to LSTM backbone. Source code is available at \url{https://github.com/YuanEZhou/cbtic}.
\end{abstract}


\vspace{-0.3cm}
\section{Introduction}

Image captioning\cite{vinyals2015show,yang2019auto,pan2020x,Zhou_2021_ICCV,huang2020image,liu2021vocabulary,yang2020auto,wang2022end,wang2023controllable,wang2023efficient,liu2023event,song2024efficiently,zhang2025deep}, which aims at describing the visual content of an image with natural language sentences, is one of the important tasks to connect vision and language.  Inspired by the sequence-to-sequence model~\cite{sutskever2014sequence} for neural machine translation, most  proposed models typically follow the encoder/decoder paradigm. In between, a convolutional neural network (CNN) or Transformer is utilized to encode an input image and recurrent neural network (RNN) or Transformer~\cite{vaswani2017attention} is adopted as sentence decoder to generate a caption. 

\begin{figure}[!t]
\centering
\includegraphics[width=2.5in]{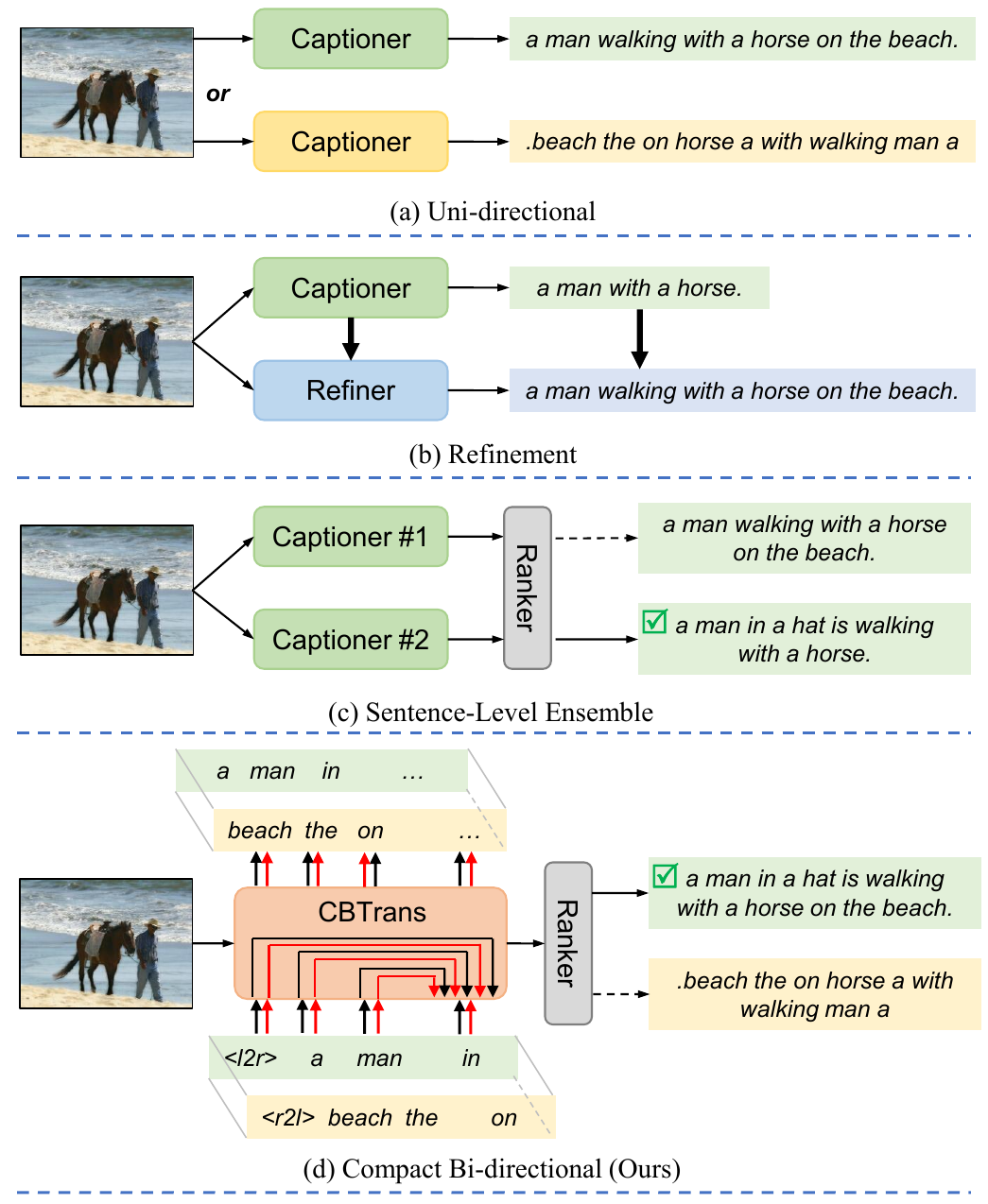}
\vspace{-0.3cm}
\caption{A conceptual overview of (a)~Uni-directional generation,including left-to-right(L2R) and right-to-left(R2L), (b)~Refinement-based generation, (c)~Sentence-Level Ensemble and (d)~our proposed Compact Bidirectional Transformer for Image Captioning~(CBTrans), which combines the advantage of (abc), where L2R and R2L 'flows' share an unified network and interaction between the two 'flows' is optionally allowed inside.}
\label{fig:figure1}
\vspace{-0.6cm}
\end{figure}

Most current image captioning models~\cite{pan2020x,zhang2021rstnet,yang2021deconfounded} typically adopt the uni-directional generation manner(as shown in Fig.~\ref{fig:figure1}(a)), which is straightforward.
During training and testing, they can only access the past context for the current prediction. This unidirectional property can't make them exploit bidirectional context  for better decoding. 

To make use of bidirectional context during decoding, refinement-based methods~\cite{Sammani2019ModificationNet,sammani2020show,wang2020show,song2021image,zhang2021open} are proposed recently. The decoder of  this type of model typically consists of two networks~(as shown in Fig.~\ref{fig:figure1}(b)). The first network is usually  a primary captioner or an image-text retriever, which is used to generate or retrieve a related sentence. After that, a senior refiner, which is the second network, generates the final caption by being allowed to attend to the sentence produced before. This semantic attention can help the senior refiner to look at both past and future semantic context  and thus improve decoding  at every time step. However, the two networks in the decoder can only be executed sequentially, which can not make full use of the parallelizability of GPU device.

In this work, we introduce a Compact Bidirectional Transformer model for image captioning~(dubbed CBTrans) that can leverage bidirectional context implicitly and explicitly while the decoder can be executed in parallel. Unlike refinement-based models that rely on two sequentially executed networks, our model is termed compact because it integrates both L2R and R2L decoding flows into a single Transformer or LSTM network with shared parameters. This design not only preserves parameter efficiency but also enables parallel bidirectional decoding within a unified architecture, allowing the decoder to exploit future context without requiring a separate refinement stage.
Specifically, it is implemented by tightly coupling the L2R and the R2L flows into a single compact model and optionally allowing explicit interaction between the two flows~(as shown in Fig.~\ref{fig:figure1}(d)). During training, each image is associated with two captions instead of a single caption as in conventional L2R captioning models.
One caption is from left to right with a $\langle l2r \rangle$ prefix and the other one is from right to left with a  $\langle r2l \rangle$  prefix. Inside the CBTrans model, the generation of a target word~(e.g., \textit{'walking'}) can not only depend on  previous words in its own flow ~(e.g.,\textit{'a man in a hat is'},i.e., past context)  but also optionally  previous words in the other flow~(e.g.,\textit{'beach the on horse a with'}, i.e., future context). The joint loss is composed of both L2R and R2L losses and the model is end-to-end trainable. During inference, CBTrans model takes both $\langle l2r \rangle$ and $\langle r2l \rangle$ as the text input in the first step  and optionally allows interaction between the two flows along the entire decoding process. Finally, the output caption of L2R 'flow' and the one of R2L 'flow' are ranked based on their  probabilities and the larger one is chosen as the output~(i.e. sentence-level ensemble).

We conduct extensive ablation studies on the MSCOCO benchmark dataset to better understand and verify the effectiveness of this model. 
We find that the \textbf{explicit} interaction slightly contributes to the final performance when a proper hyper-parameter is selected. 
We find that the \textbf{compact} architecture serves as a good regularization for \textbf{implicitly} exploiting bidirectional context  and a single CBTrans model achieves an effect of \textbf{sentence-level ensemble}, which usually needs to train and save  two models for improving final predictions as shown in Fig.~\ref{fig:figure1}(c). By combining with word-level ensemble, the effect of sentence-level ensemble is further enlarged. 
Overall, the compact bidirectional architecture and  sentence-level ensemble  play more important roles than the \textbf{explicit} interaction mechanism. We further extend the conventional one-flow self-critical training to the two-flows version under this model architecture and achieve new state-of-the-art results in comparison with non-vision-language-pretraining models. Finally, we verify the generality of this compact bidirectional architecture by extending it to LSTM backbone and proposing the CBLSTM model. 

The main contributions are summarized as follows:
\begin{itemize}
\item
We introduce a Compact Bidirectional Transformer model for image captioning that can leverage bidirectional context implicitly and explicitly while the decoder is parameter-efficient and can be executed parallelly. And we conduct extensive ablation studies to better understand this architecture.
\item
We further propose to combine word-level and sentence-level ensemble seamlessly and extend the conventional one-flow self-critical training to the two-flows version under this architecture and achieve new state-of-the-art results in comparison with non-vision-language-pretraining models. 
\item
Finally, we verify the generality of this compact bidirectional architecture
by extending it to LSTM backbone and proposing the CBLSTM model.
\end{itemize}

The remainder of this paper is organized as follows. Section~\ref{Related work} reviews related works. Section~\ref{Approach} elaborates on the proposed approach. The analysis and discussion of the experimental results are presented in Section~\ref{Experiments}. Finally, we conclude our work in Section~\ref{Conclusion}.

\section{Related work}
\label{Related work}

\subsection{Image Captioning}
Over the last few years, a broad collection of methods have been proposed in the field of image captioning. In a nutshell, we have gone through grid-feature~\cite{xu2015show,jiang2020defense} then region-feature~\cite{anderson2018bottom} and relation-aware visual feature~\cite{yao2018exploring,yang2019auto} on the image encoding side. On the sentence decoding side, we have witnessed LSTM~\cite{vinyals2015show}, CNN~\cite{gu2017empirical} and Transformer~\cite{cornia2020meshed} equipped with various attention~\cite{huang2019attention,zhou2020more,pan2020x} as the decoder. On the training side, models are typically trained by cross-entropy loss and then Reinforcement Learning~\cite{rennie2017self}, which enables the use of non-differentiable caption metrics as optimization objectives and makes a notable achievement. 
Recently, vision-language pre-training has also been adopted for image captioning and shows remarkable results. These models~\cite{zhou2020unified,li2020oscar,zhang2021vinvl} are firstly pre-trained on large image-text corpus and then finetuned.
Though impressive performance has been achieved, most state-of-the-art models adopt the left-to-right  generation  manner,which is straightforward but can't exploit bidirectional context. To make use of bidirectional context as far as possible, refinement-based methods are proposed recently, where \cite{Sammani2019ModificationNet,sammani2020show,song2021image} generate a primary caption and \cite{wang2020show,zhang2021open} retrieve a related primary caption in the first stage and both them generate a senior caption in the second stage based on the primary caption. There is also an early work~\cite{wang2016image} that tries to overcome the shortcomings of unidirectional model by combining the output captions of two separate forward and backward  LSTM networks. Different from these models, CBTrans is a single Compact Bidirectional Transformer model and CBLSTM is a single Compact Bidirectional LSTM model.

\vspace{-0.3cm}
\subsection{Multi-task Learning}
Multi-task learning is a useful learning paradigm to improve the supervision and the generalization performance of a task by jointly training it with related tasks~\cite{caruana1997multitask}. Here, we only review some multi-task learning  works related to  captioning. To the best of our knowledge, there are roughly two types so far.
One is to jointly train the task of captioning and syntax generation~(e.g. Part-of-Speech)~\cite{zhao2018multi,deshpande2019fast,wang2019controllable,hou2019joint}.
The other one is to train a unified captioning model on multilingual~(e.g. English and Chinese ) captioning datasets~\cite{elliott2015multilingual,li2019coco,wang2019vatex,song2024embedded,weng2025lgvlm}.
Different from the above multi-task learning methods, one task is the L2R generation and the other one is the R2L generation in our method.

\vspace{-0.3cm}
\subsection{Neural Machine Translation}
Neural Machine Translation~(NMT) has an important impact on the research of  captioning task. The standard encoder-decoder paradigm is derived from NMT~\cite{sutskever2014sequence, bahdanau2014neural}. A complete review is beyond the scope of this paper and we only focus on works related to endow decoder with bidirectional ability. \cite{zhang2019regularizing} trained a backward decoder jointly  with a forward decoder by matching their  output probabilities to iteratively improve each other. \cite{chen2020distilling} proposed to further encourage conventional auto-regressive decoder to plan ahead by distilling bidirectional knowledge learned in BERT~\cite{devlin2018bert}. \cite{zhang2018asynchronous} proposed asynchronous bidirectional decoding for NMT by equipping the conventional  attentional forward decoder with an auxiliary backward decoder. \cite{zhou2019synchronous} proposed a synchronous bidirectional neural machine translation that predicts its outputs using left-to-right and right-to-left decoding simultaneously and interactively. This work pursues the spirit of~\cite{zhou2019synchronous} for image captioning and we find that it is the compact bidirectional architecture and sentence-level ensemble  mechanism instead of explicit interaction that contributes more to the model. This is overlooked in~\cite{zhou2019synchronous} and indicates that explicit interaction is overestimated. What's more, we further propose to combine word-level and sentence-level ensemble seamlessly and extend the conventional one-ﬂow self-critical training to the two-ﬂows version. 


\section{Approach} 
\label{Approach}
\begin{figure*}[!t]
\centering
\includegraphics[width=0.8\textwidth]{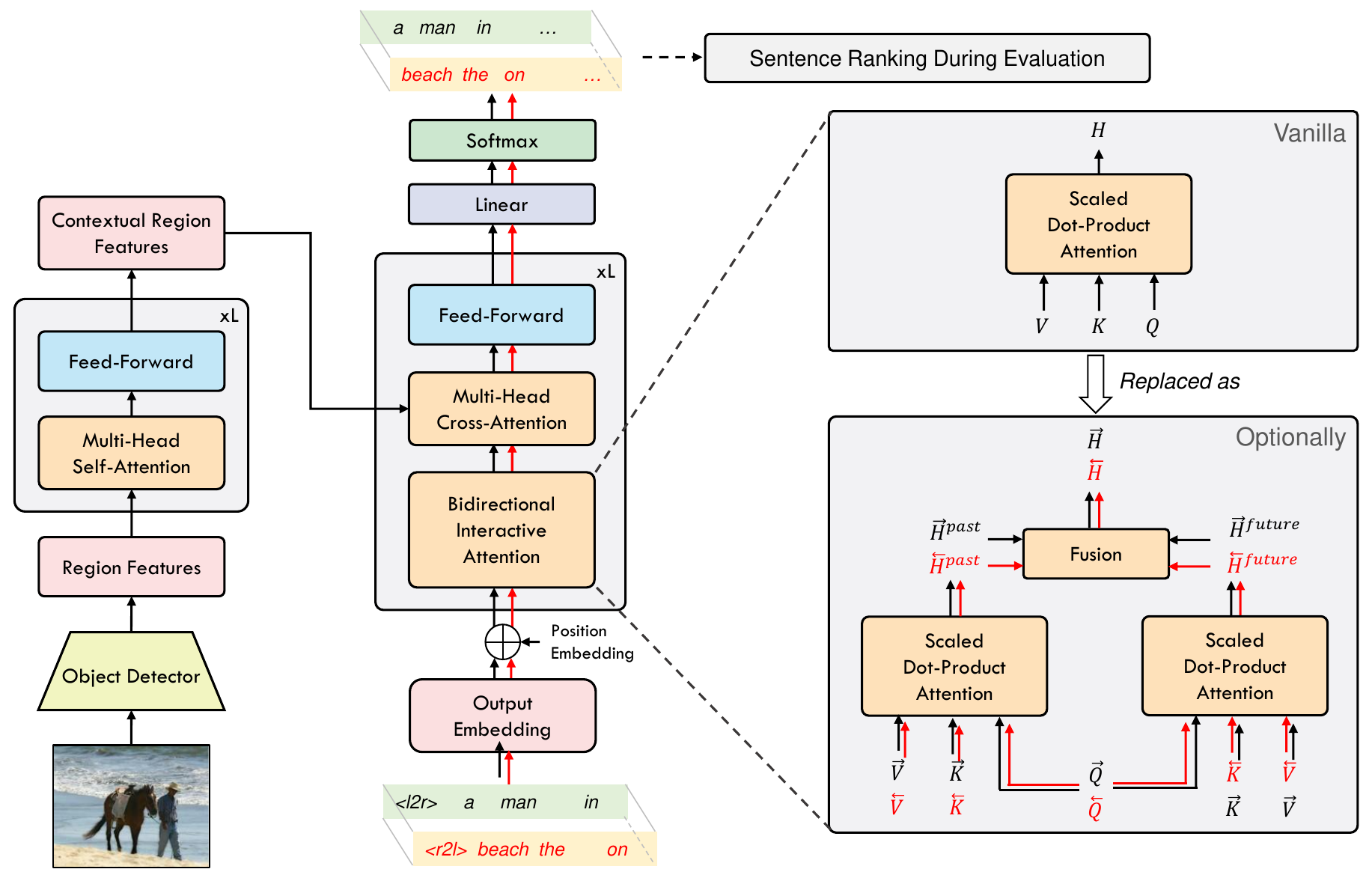}
\vspace{-0.3cm}
\caption{Illustration of Compact Bidirectional Transformer for Image Captioning~(CBTrans). CBTrans model composes of an encoder~(in the left) and a decoder~(in the middle). An intuitional illustration of the extension of the standard Scaled Dot-Product Attention module is shown on the right, enabling explicit bidirectional context interaction (see Eq.~(5)–(6)). Notice that the Residual Connections, Layer Normalization are omitted. The subscript of head $i$ in Q/K/V/H is omitted for brevity. Right-to-right~(R2L) flow is marked in red. Best view in color.}
\label{fig:framework}
\vspace{-0.6cm}
\end{figure*}


In this section, we first present two representative compact bidirectional architectures built on the well-known Transformer~\cite{vaswani2017attention} and LSTM~\cite{hochreiter1997long} and then introduce the training procedure  for model optimization. 
The compact nature of our architecture stems from the shared parameters between L2R and R2L flows, allowing bidirectional decoding within a single network, as opposed to dual-network sequential designs. There is no increase in the number of model parameters during this process.

\vspace{-0.3cm}
\subsection{CBTrans Model} 
Firstly, a pre-trained object detector~\cite{ren2015faster} represents an image \textit{I} as a  set of region features.
Then given the image region features,  CBTrans aims to take advantage of bidirectional property in a single model  to generate a sensible caption. 
The architecture of CBTrans model is illustrated  in Fig.~\ref{fig:framework}, which consists of an encoder and decoder.

\subsubsection{Image Features Encoder.} 
The encoder, which is basically the same as the Transformer encoder~\cite{vaswani2017attention},  takes the image region features as inputs and outputs the contextual region features. It consists of a stack of $L$ identical layers.  Each layer has two sublayers. The first is  a multi-head self-attention sublayer and the second is a position-wise feed-forward sublayer. Both sublayers are followed by residual connection~\cite{he2016deep} and layer normalization~\cite{ba2016layer} operations for stable training.
Multi-head self-attention builds on the scaled dot-product attention, which operates on a query Q, key K and value V as:
\vspace{-0.2cm}
\begin{equation}
    Attention(Q,K,V) = softmax(\frac{QK^T}{\sqrt{d_k}})V,
\end{equation}
where $d_k$ is the dimension of the key.  
Multi-head self-attention  firstly  projects the   queries, keys and values $h$ times with different learned linear projections and then computes scaled dot-product attention for each one. After that, it concatenates the results and projects them with another learned linear projection:
\vspace{-0.2cm}
\begin{gather}
    H_i = Attention(QW_i^Q, KW_i^K, VW_i^V), \\
    MultiHead(Q,K,V) = Concat(H_1,\dots H_h)W^O ,
\end{gather}
where $W_i^Q, W_i^K \in \mathbb{R}^{d_{model}\times d_k}$, $W_i^V \in \mathbb{R}^{d_{model}\times d_v}$ and $W^O \in \mathbb{R}^{hd_v \times d_{model}}$ . The self-attention in the encoder performs attention over itself, i.e.,  ($Q=K=V$), which is the image region features in the first layer.
After a multi-head self-attention sublayer,  the position-wise feed-forward sublayer~(FFN) is applied to each position separately and identically:
\vspace{-0.2cm}
\begin{equation}
    FFN(x) = max(0, xW_1+b_1)W_2+b_2,
\end{equation}
where $W_1\in \mathbb{R}^{d_{model}\times d_{ff}}$, $W_2\in \mathbb{R}^{d_{ff}\times d_{model}}$, $b_1\in \mathbb{R}^{d_{ff}}$ and $b_2\in\mathbb{R}^{d_{model}}$ are learnable parameter matrices.

\subsubsection{Captioning Decoder.} 
The decoder takes contextual region features and a pair of L2R and R2L word sequences of each image as input  and outputs a pair of predicted words probability sequences. To make use of order information, position encodings~\cite{vaswani2017attention} are added to word embedding features. 
It is worth emphasizing that the L2R and R2L flows in the decoder run in parallel and the explicit bidirectional interaction only optionally happens in the mask multi-head bidirectional interactive attention sublayer.
Specifically, the decoder  consists of $L$ identical layers and each layer has three sublayers:  a masked multi-head bidirectional interactive attention sublayer, a multi-head cross-attention sublayer and a position-wise feed-forward sublayer. Residual connection and layer normalization are also applied after each sublayer. The multi-head cross-attention is similar to the multi-head self-attention mentioned above except that the key and value are now contextual region features and the query is the output of its previous sublayer. 
The masked multi-head bidirectional interactive attention sublayer can be seen as an optional extension of the masked multi-head attention sublayer in the original Transformer decoder~\cite{vaswani2017attention}. The main difference between them lies in the scaled dot-product attention module. An intuitional illustration  is  shown in the right of Fig.~\ref{fig:framework}. 
Formally, the scaled dot-product attention for each head $i$ in  the original masked multi-head attention module is extended to:
\vspace{-0.2cm}
\begin{gather}
    \notag
    \overrightarrow{H}_{i}^{past}= Attention(\overrightarrow{Q}_{i},\overrightarrow{K}_{i},\overrightarrow{V}_{i})\\ 
    \overrightarrow{H}_{i}^{future}=Attention(\overrightarrow{Q}_{i},\overleftarrow{K}_{i},\overleftarrow{V}_{i}) ,\\  \notag
    \overrightarrow{H}_{i} = \overrightarrow{H}_{i}^{past}+\lambda*AF(\overrightarrow{H}_{i}^{future} )
\end{gather}

where $\overrightarrow{H}_{i}^{past}$ is the conventional one to capture past context and  $\overrightarrow{H}_{i}^{future}$ is the extended part to capture future context by using query in the L2R direction and key/value in R2L direction. $\overrightarrow{H}_{i}$ is the bidirectional-aware state of L2R direction  by non-linearly  fusing past and future context and AF denotes activation function, such as Relu or Tanh. 
It is worth noting that the self-attention in the decoder usually  is equipped  with a lower triangular  matrix mask  for preventing positions from attending to subsequent positions and we omit it here for brevity. And the bidirectional-aware state of backward direction $\overleftarrow{H}_{i}$ can be symmetrically computed as:

\begin{footnotesize}
\begin{equation}
    \overleftarrow{H}_{i} = Attention(\overleftarrow{Q}_{i},\overleftarrow{K}_{i},\overleftarrow{V}_{i})+\lambda*AF( Attention(\overleftarrow{Q}_{i},\overrightarrow{K}_{i},\overrightarrow{V}_{i})).
\end{equation}
\end{footnotesize}
In particular, the masked multi-head bidirectional interactive attention sublayer degrades to the original masked multi-head attention sublayer when $\lambda=0$.

Finally, we use a learned linear transformation and softmax function to convert the decoder output to a pair of predicted next-token probabilities $p(\overrightarrow{y}_{t}|\overrightarrow{y}_{<t},\overleftarrow{y}_{<t},I;\theta)$ and $p(\overleftarrow{y}_{t}|\overleftarrow{y}_{<t},\overrightarrow{y}_{<t},I;\theta)$, where $\overrightarrow{y}_{<t}$~($\overleftarrow{y}_{<t}$) denotes the words sequence of L2R~(R2L) direction before the $t$-th word of target $\overrightarrow{y}$~($\overleftarrow{y}$) and $\theta$ is the shared parameters. 

\vspace{-0.3cm}
\subsection{CBLSTM Model} 
\begin{figure*}[!t]
\centering
\includegraphics[width=0.8\textwidth]{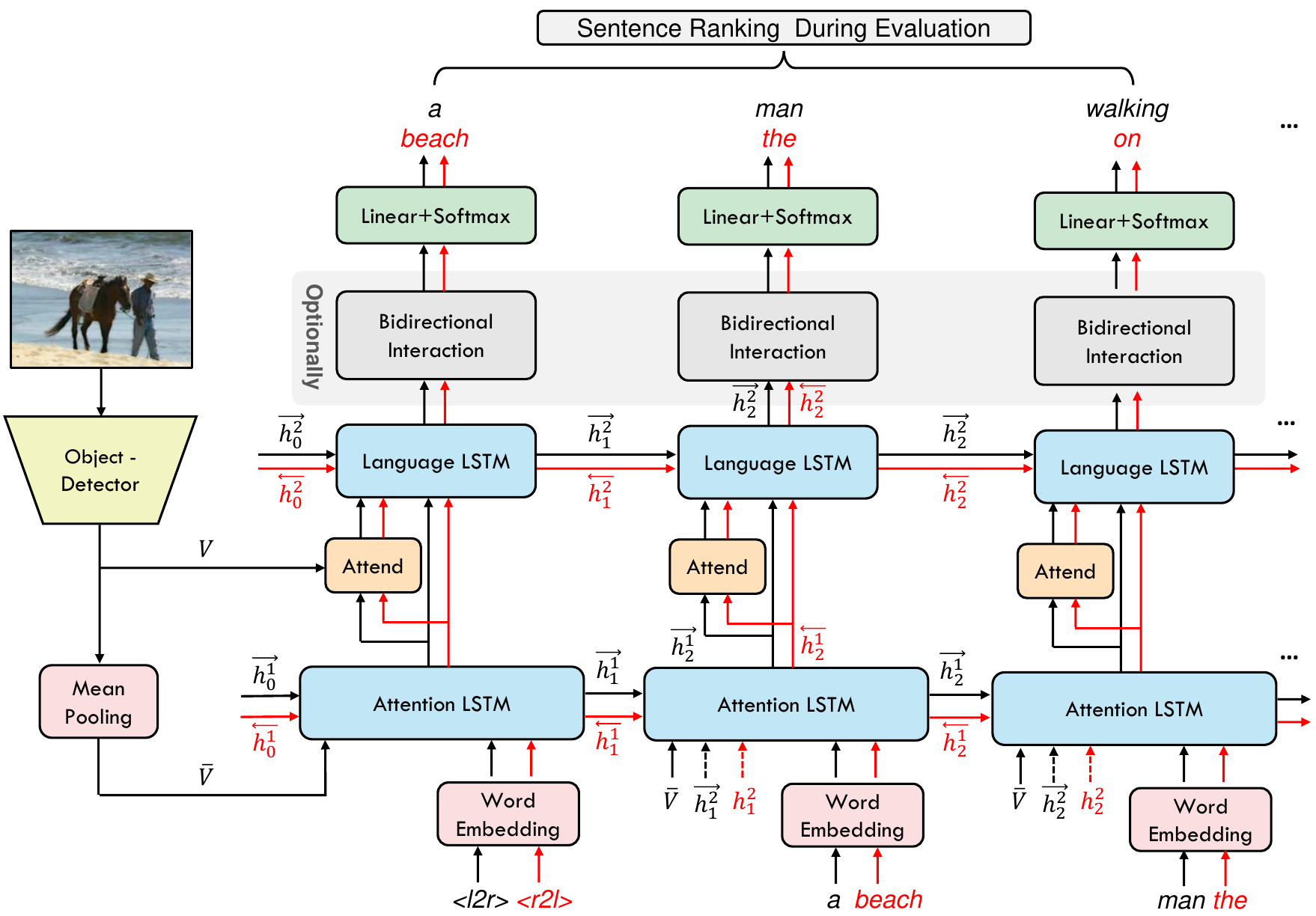}
\vspace{-0.3cm}
\caption{Illustration of Compact Bidirectional LSTM model~(CBLSTM) for Image Captioning. Best view in color.}
\label{fig:framework-CBLSTM}
\vspace{-0.6cm}
\end{figure*}

For LSTM~\cite{hochreiter1997long} backbone, we adopt the popular Up-Down~\cite{anderson2018bottom} model as our start point and our final compact bidirectional LSTM model~(dubbed as CBLSTM) is illustrated in Fig.~\ref{fig:framework-CBLSTM}, which also consists of an encoder and a decoder.

\subsubsection{Image Features Encoder} The encoder is just a vanilla object detector~\cite{ren2015faster} and followed by a feature projection matrix, which represents an image $I$ as a set of region features $V=\{v_{i}\}_{i=1}^{R}$.

\subsubsection{Captioning Decoder}
The decoder is mainly composed of two LSTM layers where the first one is the Attention LSTM and the second one is the Language LSTM. Each layer is indicated with the corresponding subscript in the following equations. Instead of having only one L2R 'flow' as in conventional Up-Down decoder~\cite{anderson2018bottom}, the decoder of our CBLSTM model processes both L2R  and  R2L flows simultaneously. The L2R flow has a $\langle l2r \rangle$ prefix and the R2L one has a $\langle r2l \rangle$ prefix.
Specifically, at time step $t$, the Attention LSTM takes previous hidden state outputs of the Language LSTM on both flows~(i.e.,$\overrightarrow{h}_{t-1}^{2}$and $\overleftarrow{h}_{t-1}^{2}$), previous word embedding of both flows~(i.e.,$\overrightarrow{e}_{t-1}=W_{e}\overrightarrow{y}_{t-1}$ and $\overleftarrow{e}_{t-1}=W_{e}\overleftarrow{y}_{t-1}$) and mean-pooled image feature $\overline{v}=\frac{1}{R}\sum_{i}v_{i}$ as input and outputs hidden state of both flows:
\vspace{-0.2cm}
\begin{gather}
\notag
    \overrightarrow{h}_{t}^{1}=LSTM_{1}([\overrightarrow{h}_{t-1}^{2};\overline{v};\overrightarrow{e}_{t-1}],\overrightarrow{h}_{t-1}^{1})\\
    \overleftarrow{h}_{t}^{1}=LSTM_{1}([\overleftarrow{h}_{t-1}^{2};\overline{v};\overleftarrow{e}_{t-1}],\overleftarrow{h}_{t-1}^{1}),
\end{gather}

where $[;]$ denotes concatenation, $W_{e}$ is the word embedding matrix and $\overrightarrow{y}_{t-1}$~( $\overleftarrow{y}_{t-1}$) is the word of L2R~(R2L) flow at time step $t-1$.
Given $\overrightarrow{h}_{t}^{1}$, the attended image feature of L2R flow  is calculated as:
\vspace{-0.2cm}
\begin{gather}
\notag
  \overrightarrow{z}_{i,t} = w_{a}^{T}tanh(W_{va}v_{i}+W_{ha}\overrightarrow{h}_{t}^{1})\\
  \boldsymbol{\overrightarrow{\beta}}_{t} = softmax(z_{t}), \quad \overrightarrow{\hat{v}_{t}} = \sum_{i=1}^{R}\overrightarrow{\beta}_{i,t}v_{i},
\end{gather}

where $w_{a}$, $W_{va}$ and $W_{ha}$ are learned weights. And $\overleftarrow{\hat{v}_{t}}$ can be calculated in the same way. Then the Language LSTM takes the attended image features~(i.e.,$\overrightarrow{\hat{v}_{t}}$ and $\overleftarrow{\hat{v}_{t}}$) and hidden state of Attention LSTM~(i.e.,$\overrightarrow{h}_{t}^{1}$ and $\overleftarrow{h}_{t}^{1}$) as input and outputs the language hidden states of both flows as:
\vspace{-0.2cm}
\begin{gather}
\notag
    \overrightarrow{h}_{t}^{2}=LSTM_{2}([\overrightarrow{h}_{t}^{1};\overrightarrow{\hat{v}_{t}}],\overrightarrow{h}_{t-1}^{2})\\
    \overleftarrow{h}_{t}^{2}=LSTM_{2}([\overleftarrow{h}_{t}^{1};\overleftarrow{\hat{v}_{t}}],\overleftarrow{h}_{t-1}^{2}).
\end{gather}

Optionally, the language hidden states can be further fed into a Bidirectional Interaction Module to obtain bidirectional context as:
\vspace{-0.2cm}
\begin{gather}
\notag
    \overrightarrow{h}_{t}^{2}=\overrightarrow{h}_{t}^{2}+ \lambda * AF(\overleftarrow{h}_{t}^{2})\\
    \overleftarrow{h}_{t}^{2}=\overleftarrow{h}_{t}^{2} + \lambda * AF(\overrightarrow{h}_{t}^{2}),
\end{gather}
where AF denotes activation function and $\lambda$ is the weight balance.

Finally, the language hidden states are converted to conditional probability distribution over possible output words as:
\begin{gather}
\notag
  p(\overrightarrow{y}_{t}|\overrightarrow{y}_{<t},\overleftarrow{y}_{<t},I;\theta) = softmax(W_{o}\overrightarrow{h}_{t}^{2}+b_{o}) \\
  p(\overleftarrow{y}_{t}|\overleftarrow{y}_{<t},\overrightarrow{y}_{<t},I;\theta) = softmax(W_{o}\overleftarrow{h}_{t}^{2}+b_{o}),
\end{gather}
 where $W_{o}$ and $b_{o}$ are learned weights and biases, $\overrightarrow{y}_{<t}$~($\overleftarrow{y}_{<t}$) denotes the words sequence of L2R~(R2L) direction before the $t$-th word and $\theta$ is the shared parameters.

\vspace{-0.3cm}
\subsection{Training} 
The whole training procedure of both models includes two stages.
At the first training stage, given a triple $(I,\overrightarrow{y},\overleftarrow{y})$,we pad the two sentences to equal length $T$ without loss of generality, i.e., $\overrightarrow{y}=(\overrightarrow{y}_{1},\overrightarrow{y}_{2},...,\overrightarrow{y}_{T})$ and $\overleftarrow{y}=(\overleftarrow{y}_{1},\overleftarrow{y}_{2},...,\overleftarrow{y}_{T})$.   We optimize the model by minimizing  joint cross-entropy~(XE) loss, which is composed of a L2R and R2L XE loss:
\vspace{-0.3cm}
\begin{equation}
\begin{aligned}
L_{XE}(\theta)=-\sum_{t=1}^{T}\{\log p(\overrightarrow{y}_{t}|\overrightarrow{y}_{<t},\overleftarrow{y}_{<t},I;\theta)
       \\  + \log p(\overleftarrow{y}_{t}|\overleftarrow{y}_{<t},\overrightarrow{y}_{<t},I;\theta)\}.
\end{aligned}
\label{eqn:sat}
\end{equation}
In order to prevent the bidirectional model from  learning to directly generate the second half of L2R flow by copying the first half of R2L flow and vice versa, $\overleftarrow{y}$ is chosen from the rest caption annotations of the same image~(each image is annotated with five captions in the dataset)  then reversed and paired with $\overrightarrow{y}$.

At the second training stage, we jointly finetune the model using  self-critical training~(SC)\cite{rennie2017self} for both L2R and R2L directions and the gradient can be expressed as:
\vspace{-0.3cm}
\begin{equation}
\begin{aligned}
    \nabla_\theta L_{SC}(\theta)=-\frac{1}{N}\sum_{n=1}^N\{(R(\overrightarrow{\hat{y}}^n)-\overrightarrow{b})\nabla_\theta\log p(\overrightarrow{\hat{y}}^n|I;  \theta) \\
    +(R(\overleftarrow{\hat{y}}^n)-\overleftarrow{b})\nabla_\theta\log p(\overleftarrow{\hat{y}}^n|I;  \theta)\},
\end{aligned}
\end{equation}
where $R$ is the CIDEr~\cite{vedantam2015cider} score function, and $b$ is the baseline score. We adopt the baseline score proposed in~\cite{luo2020better}, where the baseline score is defined as the average reward of the rest samples rather than the original greedy decoding reward. We sample $N=5$ captions for each image and each direction and $\overrightarrow{\hat{y}}^n$~($\overleftarrow{\hat{y}}^n$) is the $n$-th sampled caption of L2R~(R2L) direction.

\section{Experiments}
\label{Experiments}
\begin{table*}[ht]
    \centering
    \caption{Performance comparisons on MSCOCO Karpathy test split, where B@$N$, M, R, C and S are short for BLEU@$N$, METEOR, ROUGE-L, CIDEr and SPICE scores. All values are reported as percentage (\%). $^{\sum}$ indicates model ensemble.}
    \label{tab:offline}
    \vspace{-0.3cm}
    \resizebox{\linewidth}{!}{
    \begin{tabular}{l | c c c c c c c c | c c c c c c c c}
    \hline
		  & \multicolumn{8}{c|}{\textbf{Cross-Entropy Loss}} & \multicolumn{8}{c}{\textbf{CIDEr Score Optimization}} \\
		                             & B@1  & B@2  & B@3  & B@4  & M    & R    & C     & S    & B@1  & B@2  & B@3  & B@4  & M    & R    & C     & S  \\	
			\hline \hline
LSTM \cite{vinyals2015show}            &   -  &   -  &   -  & 29.6 & 25.2 & 52.6 & 94.0  &  -   &  -   &  -   &  -   & 31.9 & 25.5 & 54.3 & 106.3 &  -   \\
SCST \cite{rennie2017self}       &   -  &   -  &   -  & 30.0 & 25.9 & 53.4 & 99.4  &  -   &  -   &  -   &  -   & 34.2 & 26.7 & 55.7 & 114.0 &  -   \\
LSTM-A \cite{yao2017boosting}    & 75.4 &   -  &   -  & 35.2 & 26.9 & 55.8 & 108.8 & 20.0 & 78.6 &  -   &  -   & 35.5 & 27.3 & 56.8 & 118.3 & 20.8 \\
RFNet \cite{jiang2018recurrent}  & 76.4 & 60.4 & 46.6 & 35.8 & 27.4 & 56.5 & 112.5 & 20.5 & 79.1 & 63.1 & 48.4 & 36.5 & 27.7 & 57.3 & 121.9 & 21.2 \\
Up-Down \cite{anderson2018bottom}& 77.2 &   -  &   -  & 36.2 & 27.0 & 56.4 & 113.5 & 20.3 & 79.8 &  -   &  -   & 36.3 & 27.7 & 56.9 & 120.1 & 21.4 \\
GCN-LSTM \cite{yao2018exploring} & 77.3 &   -  &   -  & 36.8 & 27.9 & 57.0 & 116.3 & 20.9 & 80.5 &  -   &  -   & 38.2 & 28.5 & 58.3 & 127.6 & 22.0 \\
LBPF \cite{qin2019look}           & 77.8 &   -  &   -  & 37.4 & 28.1 & 57.5 & 116.4 & 21.2 & 80.5 &  -   &  -   & 38.3 & 28.5 & 58.4 & 127.6 & 22.0 \\
SGAE \cite{yang2019auto}          & 77.6 &   -  &   -  & 36.9 & 27.7 & 57.2 & 116.7 & 20.9 & 80.8 &  -   &  -   & 38.4 & 28.4 & 58.6 & 127.8 & 22.1 \\
AoANet \cite{huang2019attention}  & 77.4 &   -  &   -  & 37.2 & 28.4 & 57.5 & 119.8 & 21.3 & 80.2 &  -   &  -   & 38.9 & 29.2 & 58.8 & 129.8 & 22.4 \\
X-LAN \cite{pan2020x}                        & 78.0 & \textbf{62.3} & \textbf{48.9} & \textbf{38.2} & 28.8 & 58.0 & 122.0 & 21.9 & 80.8 & 65.6 & 51.4 & 39.5 & 29.5 & 59.2 & 132.0 & 23.4 \\
M$^2$ \cite{cornia2020meshed}                       & - & - & - & - & - & - & - & - & 80.8 & - & - & 39.1 & 29.2 & 58.6 & 131.2 & 22.6 \\
X-Transformer \cite{pan2020x}                         & 77.3 & 61.5 & 47.8 & 37.0 & 28.7 & 57.5 & 120.0 & 21.8 & 80.9 & 65.8 & 51.5 & 39.7 & 29.5 & 59.1 & 132.8 & 23.4 \\
RSTNet \cite{zhang2021rstnet}                        & - & - & - & - & - & - & - & - & 81.8 & - & - & 40.1 & 29.8 & 59.5 & 135.6 & 23.3 \\
SGT \cite{li2023modeling} & 78.0 & - & - & 38.0 & 28.8 &58.1 &120.8  & - &81.4 & - & - &39.8 & 29.6 &59.2  & 132.9  & -\\
MD-SAN \cite{ji2022multi} & - & - & - & - & - & -  & -  & - 
&81.5 & - & - &39.8 &29.6 &59.1 &135.1 &23.2\\
FeiM \cite{yan2024exploring} &\textbf{78.6} & - & - &38.1  &28.8  &58.3  &122.5  &21.8  
&\textbf{81.9} & - & - &\textbf{40.5} &\textbf{29.9} &\textbf{59.8} &135.2 &23.7\\
ACfiC \cite{cai2025toward} &77.8 & 62.2 & 48.4 &37.6  &28.6  &\textbf{58.7}  &\textbf{122.7}  &21.9 &81.3 & \textbf{66.6} & \textbf{52.4} &39.6 &29.4 &59.4 &134.2 &23.3 \\
I$^2$OA \cite{zhang2025intra} & 76.5 & - & - & 36.9 & 28.9 & 57.8  & 121.3  & 22.1
&\textbf{81.9} & - & - &\textbf{40.5} &\textbf{29.9} &59.6 &136.2 &23.5
\\
\hline
CBLSTM                         & 77.6 & 62.0 & 48.3 & 37.5 & 28.9 & 57.9 & 121.0 & 22.0 & 80.7 & 65.3 & 50.4 & 38.1 & 29.0 & 58.2 & 133.6 & 22.8 \\
CBTrans                         & 78.0 & 62.2 & 48.5 & 37.5 & \textbf{29.1} & 58.2 & 122.6 & \textbf{22.3} & 81.4 & 66.5 & 51.9 & 39.6 & \textbf{29.9} & 59.1 & \textbf{136.9} & \textbf{24.0} \\\hline

			& \multicolumn{16}{c}{\textbf{Ensemble}} \\ \hline
SCST \cite{rennie2017self}$^{\sum}$ & -   &   -  &   -  & 32.8 & 26.7 & 55.1 & 106.5 &   -  &  -   &  -   &  -   & 35.4 & 27.1 & 56.6 & 117.5 &  -   \\
RFNet \cite{jiang2018recurrent}$^{\sum}$ & 77.4 & 61.6 & 47.9 & 37.0 & 27.9 & 57.3 & 116.3 & 20.8 & 80.4 & 64.7 & 50.0 & 37.9 & 28.3 & 58.3 & 125.7 & 21.7 \\
GCN-LSTM \cite{yao2018exploring}$^{\sum}$& 77.4 & - & - & 37.1 & 28.1 & 57.2 & 117.1 & 21.1 & 80.9 &  -   &  -   & 38.3 & 28.6 & 58.5 & 128.7 & 22.1 \\
SGAE \cite{yang2019auto}$^{\sum}$   &  -   &   -  &   -  &  -   &   -  &   -  &   -   &   -  & 81.0 &  -   &  -   & 39.0 & 28.4 & 58.9 & 129.1 & 22.2 \\
HIP \cite{yao2019hierarchy}$^{\sum}$ & -   &   -  &   - &  {38.0}  &  {28.6}  &  {57.8}  &  {120.3} & {21.4} & -   &   -  &   -  &  {39.1}  &  {28.9}  &  {59.2}  & {130.6}   & {22.3} \\
AoANet \cite{huang2019attention}$^{\sum}$ & 78.7 & - & - & 38.1 & 28.5 & 58.2 & 122.7 & 21.7 & 81.6 &  -   &  -   & 40.2 & 29.3 & 59.4 & 132.0 & 22.8 \\
M$^2$ T\cite{cornia2020meshed}$^{\sum}$ & \textbf{-} & \textbf{-} & \textbf{-} & \textbf{-} & \textbf{-} & \textbf{-} & \textbf{-} & \textbf{-} & 82.0 & - & - & 40.5 & 29.7 & 59.5 & 134.5 & 23.5 \\
X-LAN\cite{pan2020x}$^{\sum}$ & 78.8 & 63.4 & 49.9 & 39.1 & 29.1 & 58.5 & 124.5 & 22.2 & 81.6 & 66.6 & 52.3 & 40.3 & 29.8 & 59.6 & 133.7 & 23.6 \\
X-Transformer\cite{pan2020x}$^{\sum}$ & 77.8 & 62.1 & 48.6 & 37.7 & 29.0 & 58.0 & 122.1 & 21.9 & 81.7 & 66.8 & 52.6 & 40.7 & 29.9 & 59.7 & 135.3 & 23.8 \\ \hline
CBLSTM & 78.3 & 62.8 & 49.1 & 38.2 & 29.2 & 58.5 & 123.1  & 22.2 & 81.5 & 66.3 & 51.5 & 39.1 & 29.3 & 58.8 & 135.6 & 23.0 \\
CBTrans & \textbf{79.1} & \textbf{63.9} & \textbf{50.3} & \textbf{39.4} & \textbf{29.8} & \textbf{59.3} & \textbf{127.8} & \textbf{23.1} & \textbf{82.7} & \textbf{67.9} & \textbf{53.5} & \textbf{41.1} & \textbf{30.4} & \textbf{60.1} & \textbf{140.3} & \textbf{24.3} \\
\hline
	
    \end{tabular}
     }
     \vspace{-0.3cm}
\end{table*}

\begin{table}[ht]
\small
\centering
\caption{Performance of CBTrans on MSCOCO validation set with different activation functions and $\lambda$. Only cross-entropy loss is used and beam size is set to 1.}
\vspace{-0.3cm}
\label{tab:lambda}
\resizebox{\linewidth}{!}{
\begin{tabular}{l|ccccccccccccc}
\toprule
  \multirow{2}{*}{\diagbox{AF}{Metrics}{$\lambda$}}& & \multicolumn{2}{c}{0.0} & & \multicolumn{2}{c}{0.1} & & \multicolumn{2}{c}{0.4}   \\
\cmidrule{3-4} \cmidrule{6-7} \cmidrule{9-10}  
& & B@1 & C && B@1 & C & & B@1 & C   \\
\midrule
\multicolumn{10}{c}{Up-Down Feat.}  \\ 
\midrule
Tanh & & 76.5 & 114.2 && 76.5 & 114.0 & & 76.8 & 114.4  \\
ReLU & & 76.5 & 114.2 & & \textbf{76.9} & \textbf{115.0} & & 76.6 & 114.4  \\
\midrule
\multicolumn{10}{c}{VinVL Feat.}  \\ 
\midrule
ReLU & & 78.1 & 121.6 & & \textbf{78.5} & \textbf{122.4} & & 78.3 & 121.5 \\
\bottomrule
\end{tabular}
}
\vspace{-0.3cm}
\end{table}

\begin{table}[ht]
\begin{center}
\caption{Performance of CBLSTM on MSCOCO validation set with different $\lambda$. Beam size is set to 1 and ReLU activation function is selected as default.}
\vspace{-0.3cm}
\label{tab:CBLSTM-lambda}
\begin{tabular}{c|cccc}
\toprule
\multirow{2}{*}{\diagbox{$\lambda$}{Metrics}} & \multicolumn{4}{c}{Cross-Entropy Loss} \\
& B@4 & M & R & C\\ \midrule
0.0 & 37.3 & 28.7 & 58.1 & 120.6 \\
0.1 & 37.4 & 28.8 & 58.1 & 120.8\\
0.2 & 37.6 & 28.7 & 58.3 & 121.4 \\
0.4 & \textbf{37.7} & \textbf{28.9} & \textbf{58.4} & \textbf{121.4} \\
\bottomrule
\end{tabular}
\end{center}
\vspace{-0.3cm}
\end{table}

\subsection{Dataset and Evaluation Metrics.} MSCOCO~\cite{chen2015microsoft} 
is the widely used  benchmark for image captioning. 
We use the 'Karpathy' splits~\cite{karpathy2015deep} for offline experiments. This split contains $113,287$ training images, $5,000$ validation images and $5,000$ testing images. Each image has $5$ captions.
We also upload generated captions of MSCOCO official testing set, which contains $40,775$ images for online evaluation.
The quality of captions is evaluated by standard metrics\cite{chen2015microsoft}, including BLEU-1/4~\cite{papineni2002bleu}, METEOR~\cite{banerjee2005meteor}, ROUGE~\cite{lin2004rouge}, SPICE~\cite{anderson2016spice}, and CIDEr~ \cite{vedantam2015cider}, denoted as B1/4, M, R, S, and C for short.

\subsection{Implementation Details.}
Each image is represented as a set of region features with $2,048$ dimensions extracted by object detector~\cite{ren2015faster}. We try two kinds of features,i.e.,Up-Down feature with $36$ regions per image~\cite{anderson2018bottom},and VinVL feature with up to $50$ regions per image~\cite{zhang2021vinvl}, which is extracted by a more powerful detector. 
The dictionary is built by dropping the words that occur less than $5$ times and ends up with a vocabulary of $9,487$. Captions longer than $16$ words are truncated. 
Our CBTrans model almost follows the same model hyper-parameters setting as in  \cite{vaswani2017attention}
~($d_{model} = 512, d_k = d_v = 64, d_{ff} = 2048, L = 6,h = 8, p_{dropout} = 0.1$). And the word embedding size, LSTM hidden state size and image feature projection size of our CBLSTM model are all set to $1,024$.
As for the training process, we train CBTrans~(CBLSTM) under cross-entropy loss for $15$~($30$) epochs with a mini batch size of $10$~($32$). 

The optimizer in ~\cite{vaswani2017attention} is used with a learning rate initialized by 5e-4 and the warmup step is set to $20,000$ for CBTrans model and Adam~\cite{kingma2014adam} optimizer is used with learning rate initialized by 5e-4 and decayed by a factor 0.8 every three epochs for CBLSTM model.  We increase the scheduled sampling probability by 0.05 every 5 epochs.
We then optimize the CIDEr score with self-critical training for another 15~(20) epochs with an initial learning rate of 1e-5~(5e-5) for CBTrans~(CBLSTM) model.
During testing, unless stated otherwise, we use the standard beam search for conventional Transformer and Up-Down models with beam size $3$, which is the best width based on our observation. It is worth noting that the beam search method for CBTrans and CBLSTM models is a little different from the standard one. Specifically, if the total beam size is $4$, then both L2R and R2L flows should independently keep a standard beam search with beam size $2$ for each step. Unless stated otherwise, the beam size of each flow is set to $2$ for CBTrans and CBLSTM models.

It is worth noting that during beam search for CBTrans/CBLSTM, the L2R and R2L flows decode in lockstep. At each timestep, both flows generate one token based on their own past and the other flow's past. If one flow finishes earlier (predicts an $\langle end \rangle$ token), its sequence remains static for subsequent steps while the other continues, allowing the longer flow to still leverage the completed ‘future context’ from the finished flow via the bidirectional attention mechanism.

\begin{table}[ht]
\begin{center}
\caption{Ablation studies of CBTrans model on MSCOCO validation set. The default random seed is $0$.}
\vspace{-0.3cm}
\label{tab:ablation-study-CBTrans}
\resizebox{\linewidth}{!}{
\begin{tabular}{ll|cccc}
\toprule
\multirow{2}{*}{} & \multirow{2}{*}{\diagbox{Models}{Metrics}} &  \multicolumn{4}{c}{Cross-Entropy Loss} \\
&   & B@4 & M & R & C \\ \midrule
\multicolumn{6}{c}{Up-Down Feat.}  \\
\cmidrule{1-6}
1 & Transformer(L2R) & 35.4 & 27.8 & 56.3 & 111.7 \\
2 & Transformer(L2R-seed1) & 35.4 & 27.7 & 56.3 & 112.3 \\
3 & Sentence-Level Ensemble & 35.1 & 28.1 & 56.4  & 112.5 \\ \midrule
4 & Transformer(L2R) & 35.4 & 27.8 & 56.3 & 111.7 \\
5 & Transformer(R2L) & 33.9 & 27.1 & 55.2 & 109.6 \\
6 & Sentence-Level Ensemble & 34.6 & 27.8 & 56.0 & 111.7 \\ \midrule
7 & CBTrans(only keep R2L during eval.) & 34.3 & 27.5 & 56.0 & 112.6 \\
8 & CBTrans(only keep L2R during eval.) & \textbf{35.7} & 27.7 & 56.7  & 113.4\\
9 & CBTrans & 35.6 & \textbf{28.1} & \textbf{56.8} & \textbf{114.4} \\ \midrule
\multicolumn{5}{c}{VinVL Feat.}  \\ 
\cmidrule{1-6}
10 & Transformer(L2R) & 37.1 & 28.4 & 57.3 & 117.6 \\
11 & Transformer(L2R-seed1) & 36.6 & 28.6 & 57.2 & 117.3 \\
12 & Sentence-Level Ensemble & 36.8 & 28.8 & 57.4  & 118.5 \\ \midrule
13 & Transformer(L2R) & 37.1 & 28.4 & 57.3 & 117.6 \\
14 & Transformer(R2L) & 35.5 & 28.1 & 56.9 & 117.0 \\
15 & Sentence-Level Ensemble & 37.1 & 29.0 & 57.9 & 119.9 \\ \midrule
16 & CBTrans(only keep R2L during eval.) & 36.1 & 28.4 & 57.3 & 119.4 \\
17 & CBTrans(only keep L2R during eval.) & 37.8 & 28.9 & 58.0  & 121.3 \\ 
18 & CBTrans & \textbf{37.8} &  \textbf{29.1} & \textbf{58.2} & \textbf{121.8} \\ \midrule
19 & Transformer(L2R+L2R) & 37.1 &  28.5 & 57.4 & 119.0 \\

\bottomrule
\end{tabular}}
\end{center}
\vspace{-0.4cm}	
\end{table}

\subsection{Quantitative Results}

In this section, we will quantitatively analyze our models in detail by answering the following questions.

\subsubsection{What is the effect of Fusion function with different $\lambda$?}
Following previous practice~\cite{zhou2019synchronous}, we choose to  use the simple non-linear fusion mechanism. From Table~\ref{tab:lambda}, we can find that the explicit interaction mechanism only improves slightly over the baseline~($\lambda=0$) when proper $\lambda=0.1$ is selected. This indicates that the explicit interaction mechanism is not the main contributor of this model, which is different from ~\cite{zhou2019synchronous}, and motivates us to further decipher this architecture. In the following, we select ReLU and set $\lambda=0.1$ for CBTrans model as default, unless stated otherwise. We also observe similar results for CBLSTM model as shown in Table~\ref{tab:CBLSTM-lambda}. As an extreme try, we set $\lambda=0$ for CBLSTM model in the following, which means we discard the explicit Bidirectional Interaction module.

\begin{table}[ht]
\begin{center}
\vspace{-0.3cm}
\caption{Performance of CBTrans model combined with model ensemble on MSCOCO validation set. '$*2$' indicates we average the word-level output probability distributions of two independently trained instances with different parameter initialization. Beam size is set to $1$ and VinVL feature is used.}
\label{tab:ablation-study-ensemble-CBTrans}
\resizebox{\linewidth}{!}{
\begin{tabular}{l|cccc}
\toprule
\multirow{2}{*}{\diagbox{Models}{Metrics}} &  \multicolumn{4}{c}{Cross-Entropy Loss} \\
& B@4 & M & R & C \\ \midrule
CBTrans*2 & 38.3 & 29.4 & 58.9 & 125.3 \\
CBTrans(only keep L2R during eval.)*2 & 37.4 & 28.9 & 58.3 & 122.5 \\
\midrule
CBTrans*3 & 39.2 & 29.7 & 59.2 & 127.4 \\
CBTrans(only keep L2R during eval.)*3 & 38.3 & 29.3 & 58.8 & 125.1 \\\midrule
CBTrans*4 & 39.1 & 29.7 & 59.3 & 127.6 \\
CBTrans(only keep L2R during eval.)*4 & 38.6 & 29.3 & 59.0 & 125.5 \\
\bottomrule
\end{tabular}
}
\end{center}
\vspace{-0.6cm}	
\end{table}

\begin{table}[ht]
\begin{center}
\caption{Performance of CBLSTM model combined with model ensemble on MSCOCO validation set. '$*2$' indicates we average the word-level output probability distributions of two independently trained instances with different parameter initialization. Beam size is set to $1$ and VinVL feature is used.}
\vspace{-0.3cm}
\label{tab:ablation-study-ensemble-CBLSTM}
\resizebox{\linewidth}{!}{
\begin{tabular}{l|cccc}
\toprule
\multirow{2}{*}{\diagbox{Models}{Metrics}} &  \multicolumn{4}{c}{Cross-Entropy Loss} \\
& B@4 & M & R & C \\ \midrule
CBLSTM*2 & 37.7 & 28.9 & 58.4 & 121.5 \\
CBLSTM(only keep L2R during eval.)*2 & 36.5 & 28.2 & 57.6 & 117.3 \\
\midrule
CBLSTM*3 & 38.1 & 29.0 & 58.7 & 122.7 \\
CBLSTM(only keep L2R during eval.)*3 & 36.7 & 28.3 & 57.8 & 117.7 \\\midrule
CBLSTM*4 & 38.4 & 29.1 & 58.7 & 123.3 \\
CBLSTM(only keep L2R during eval.)*4 & 36.6 & 28.4 & 57.8 & 118.1 \\
\bottomrule
\end{tabular}
}
\end{center}
\vspace{-0.3cm}	
\end{table}

\subsubsection{What is the effect of each component that constitutes the compact bidirectional architecture?}
To further investigate the effectiveness of our models and each component, we conduct substantial ablation studies, as shown in Table~\ref{tab:ablation-study-CBTrans}.
Before diving into the details, we emphasize that our model doesn't increase any model parameters except two special symbols~(i.e., $\langle l2r \rangle$ and $\langle r2l \rangle$) embedding compared to the standard baseline and they share the same training script.

Firstly, by comparing the 9th row with the 1st and 5th rows, we can find that the whole CBTrans model outperforms Transformer(L2R) and Transformer(R2L) models more than 2.7\% in CIDEr metric.
And the advantage of our CBTrans model is more obvious when using better VinVL feature by comparing the 18th row with the 10th row~(e.g.,more than 4.2\% gain in CIDEr metric). 
This proves the overall effectiveness of CBTrans model. We can also see similar performance gains~(e.g., about 2.9\% gain in CIDEr metric ) for CBLSTM model by comparing the 3rd row with the 1st row in Table~\ref{tab:ablation-study-updown}.

\begin{table*}[ht]
\centering
\caption{Leaderboard of the published state-of-the-art image captioning models on the MSCOCO online testing server, where B@$N$, M, R, and C are short for BLEU@$N$, METEOR, ROUGE-L, and CIDEr scores. All values are reported as percentage (\%).}
\label{tab:online}
\resizebox{\linewidth}{!}{
\begin{tabular}{lccccccccccccccccccccc}
\toprule
  \multirow{2}{*}{\diagbox{Model}{Metrics}}& & \multicolumn{2}{c}{BLEU-1} & & \multicolumn{2}{c}{BLEU-2} & & \multicolumn{2}{c}{BLEU-3} & & \multicolumn{2}{c}{BLEU-4} & & \multicolumn{2}{c}{METEOR} & &  \multicolumn{2}{c}{ROUGE} & & \multicolumn{2}{c}{CIDEr} \\
\cmidrule{3-4} \cmidrule{6-7} \cmidrule{9-10} \cmidrule{12-13} \cmidrule{15-16} \cmidrule{18-19} \cmidrule{21-22}
& & c5 & c40 & & c5 & c40 & & c5 & c40 & & c5 & c40 & & c5 & c40 & & c5 & c40 & & c5 & c40 \\
\midrule
SCST~\cite{rennie2017self} & & 78.1 & 93.7 & & 61.9 & 86.0 & & 47.0 & 75.9 & & 35.2 & 64.5 & & 27.0 & 35.5 & & 56.3 & 70.7 & & 114.7 & 116.7 \\
Up-Down~\cite{anderson2018bottom} & & 80.2 & 95.2 & & 64.1 & 88.8 & & 49.1 & 79.4 & & 36.9 & 68.5 & & 27.6 & 36.7 & & 57.1 & 72.4 & & 117.9 & 120.5 \\
RFNet~\cite{jiang2018recurrent} & & 80.4 & 95.0 & & 64.9 & 89.3 & & 50.1 & 80.1 & & 38.0 & 69.2 & & 28.2 & 37.2 & & 58.2 & 73.1 & & 122.9 & 125.1\\
GCN-LSTM~\cite{yao2018exploring} & & 80.8 & 95.9 & & 65.5 & 89.3 & & 50.8 & 80.3 & & 38.7 & 69.7 & & 28.5 & 37.6 & & 58.5 & 73.4 & & 125.3 & 126.5 \\
SGAE~\cite{yang2019auto} & & 81.0 & 95.3 & & 65.6 & 89.5 & & 50.7 & 80.4 & & 38.5 & 69.7 & & 28.2 & 37.2 & & 58.6 & 73.6 & & 123.8 & 126.5 \\
ETA~\cite{li2019entangled} & & 81.2 & 95.0 & & 65.5 & 89.0 & & 50.9 & 80.4 & & 38.9 & 70.2 & & 28.6 & 38.0 & & 58.6 & 73.9 & & 122.1 & 124.4 \\
AoANet~\cite{huang2019attention} & & 81.0 & 95.0 & & 65.8 & 89.6 & & 51.4 & 81.3 & & 39.4 & 71.2 & & 29.1 & 38.5 & & 58.9 & 74.5 & & 126.9 & 129.6 \\
GCN-LSTM+HIP~\cite{yao2019hierarchy} & & 81.6 & 95.9 & & 66.2 & 90.4 & & 51.5 & 81.6 & & 39.3 & 71.0 & & 28.8 & 38.1 & & 59.0 & 74.1 & & 127.9 & 130.2 \\
$M^{2}$ T \cite{cornia2020meshed} & & 81.6 & 96.0 & & 66.4 & 90.8 & & 51.8 & 82.7 & & 39.7 & 72.8 & & 29.4 & 39.0 & & 59.2 & 74.8 & & 129.3 & 132.1 \\
X-Transformer\cite{pan2020x} & & 81.9 & 95.7 & & 66.9 & 90.5 & & 52.4 & 82.5 & & 40.3 & 72.4 & & 29.6 & 39.2 & & 59.5 & 75.0 & & 131.1 & 133.5 \\
GET~\cite{ji2021improving} & & 81.6 & 96.1 & & 66.5 & 90.9 & & 51.9 & 82.8 & & 39.7 & 72.9 & & 29.4 & 38.8 & & 59.1 & 74.4 & & 130.3 & 132.5 \\
RSTNet~\cite{zhang2021rstnet} & & 82.1 & 96.4 & & 67.0 & 91.3 & & 52.2 & 83.0 & & 40.0 & 73.1 & & 29.6 & 39.1 & & 59.5 & 74.6 & & 131.9 & 134.0 \\
MD-SAN \cite{ji2022multi} & & \textbf{82.4} & 96.5 & & 67.4 & 91.6 & & \textbf{52.8}  & 83.6 & & \textbf{40.7} & 73.7  & & 29.8 & 39.4 & & \textbf{59.8} & 75.0 & & 133.4 & 135.4 \\
I$^2$OA \cite{zhang2025intra} & & \textbf{82.4} & \textbf{96.6} & & 67.4 & 91.6 & & \textbf{52.8} & \textbf{83.8} & & 40.6 & \textbf{74.0}  & & 29.9 & 39.5 & & \textbf{59.8} & \textbf{75.3} & & 133.7 & 135.6 \\
\midrule
CBTrans & & 82.3 & 96.5 & & \textbf{67.5} & \textbf{91.8} & & \textbf{52.8} & 83.7 & & 40.5 & 73.5 & & \textbf{30.0} & \textbf{39.6} & & 59.6 & 75.0 & & \textbf{135.0} & \textbf{138.6} \\
\bottomrule
\end{tabular}
}
\vspace{-0.3cm}
\end{table*}

\begin{table}[ht]
\small
\begin{center}
\vspace{-0.3cm}	
\caption{Performance comparison with Transformer based alternatives, which have the same parameters, on MSCOCO validation set.}
\vspace{-0.3cm}	
\label{tab:comparison-CBTrans}
\begin{tabular}{l|cccc}
\toprule
\multirow{2}{*}{\diagbox{Models}{Metrics}} & \multicolumn{4}{c}{CIDEr Score Optimization} \\
& B@4 & M & C & S\\ \midrule
\multicolumn{5}{c}{Up-Down Feat.}  \\
\cmidrule{1-5}
Transformer(L2R) & \textbf{38.9} & 28.9 & 127.6 & 22.6 \\
Transformer(R2L) & 36.5 & 28.9 & 129.2 & 22.7 \\
CBTrans & 38.0 & \textbf{29.1} & \textbf{129.8} & \textbf{22.8} \\
\midrule
\multicolumn{5}{c}{VinVL Feat.}  \\ 
\cmidrule{1-5}
Transformer(L2R) & \textbf{41.0} & 29.8 & 134.9 & 23.6 \\
Transformer(R2L) & 37.8 & 29.6 & 136.1 & 23.6 \\
CBTrans & 40.0 & \textbf{30.1} & \textbf{137.5} & \textbf{24.2} \\
\bottomrule
\end{tabular}
\end{center}
\vspace{-0.6cm}	
\end{table}

\begin{table}[ht] 
\begin{center}
\caption{Ablation studies of CBLSTM model on MSCOCO validation set. VinVL feature is used here. The default random seed is $0$.}
\label{tab:ablation-study-updown}
\small
\begin{tabular}{@{}ll|cccc@{}}
\toprule
 & \multirow{2}{*}{\diagbox{Models}{Metrics}} &  \multicolumn{4}{c}{Cross-Entropy Loss} \\
&   & B@4 & M & R & C \\ \midrule
 & UpDown(L2R) & 37.8 & 28.5 & 57.7 & 118.5 \\
 & UpDown(R2L) & 36.8 & 28.3 & 57.5 & 117.8 \\
  & UpDown(L2R+L2R) & 37.6 & 28.6 & 57.8 & 119.3 \\
 & CBLSTM & \textbf{38.3} &  \textbf{29.0} & \textbf{58.2} & \textbf{121.4} \\

\bottomrule
\end{tabular}
\end{center}
 \vspace{-0.4cm}	
\end{table}

\begin{table}[ht] 
\small
\begin{center}
\caption{Performance comparison with up-down based alternatives, which have the same parameters, on MSCOCO validation set. VinVL feature is used here.}
\label{tab:comparison-CBLSTM}
\begin{tabular}{l|cccc}
\toprule
\multirow{2}{*}{\diagbox{Models}{Metrics}} & \multicolumn{4}{c}{CIDEr Score Optimization} \\
& B@4 & M & C & S\\ \midrule
UpDown(L2R) & 39.0 & 29.0 & 130.6 & 22.5 \\
UpDown(R2L) & 36.1 & 28.7 & 132.0 & 22.5 \\
CBLSTM & 38.4 & \textbf{29.1} & \textbf{133.5} & \textbf{22.9} \\
\midrule
\multicolumn{5}{c}{Removing Bad Endings}  \\
\cmidrule{1-5}
UpDown(R2L) & 37.5 & 28.6 & 130.8 & 22.4 \\
CBLSTM & \textbf{39.1} & \textbf{29.1} & 132.9 & 22.8 \\
\bottomrule
\end{tabular}
\end{center}
\vspace{-0.4cm}	
\end{table}

Secondly, by comparing the 7th/8th row with the 5th/1st row in  Table~\ref{tab:ablation-study-CBTrans}, we can find that the \textbf{compact} architecture of CBTrans model can serve as a good regularization by improving the performance from 109.6/111.7 CIDEr to 112.6/113.4 CIDEr. And we can also see similar trend when using better VinVL feature by comparing the 16th/17th row with the 14th/13th row. However, the regularization effect is not obvious as for CBLSTM model. One possible reason is that the model capacity of CBLSTM is very limited compared to CBTrans model.

Thirdly, we investigate the effect of sentence-level ensemble, which is a natural part of our compact bidirectional models since we have to choose one final caption from the output of  L2R and R2L 'flow'. By comparing the 9th row with the 8th row of Table~\ref{tab:ablation-study-CBTrans}, we can see the effect of sentence-level ensemble mechanism by improving the METEOR metric from $27.7$ to $28.1$. The effect of sentence-level ensemble is further enlarged when combined with model ensemble, which typically averages the word-level output probability distributions of multiple independently trained instances with different parameter initialization. From Table~\ref{tab:ablation-study-ensemble-CBTrans}, we can see that sentence-level ensemble contributes more than $2$\% in CIDEr metric at all model ensemble cases for CBTrans model. As for CBLSTM model, the sentence-level ensemble contributes more than $4$\% in CIDEr metric at all model ensemble cases, as shown in Table~\ref{tab:ablation-study-ensemble-CBLSTM}.

\begin{figure*}[ht]
	\centering
	\includegraphics[width=0.9\textwidth]{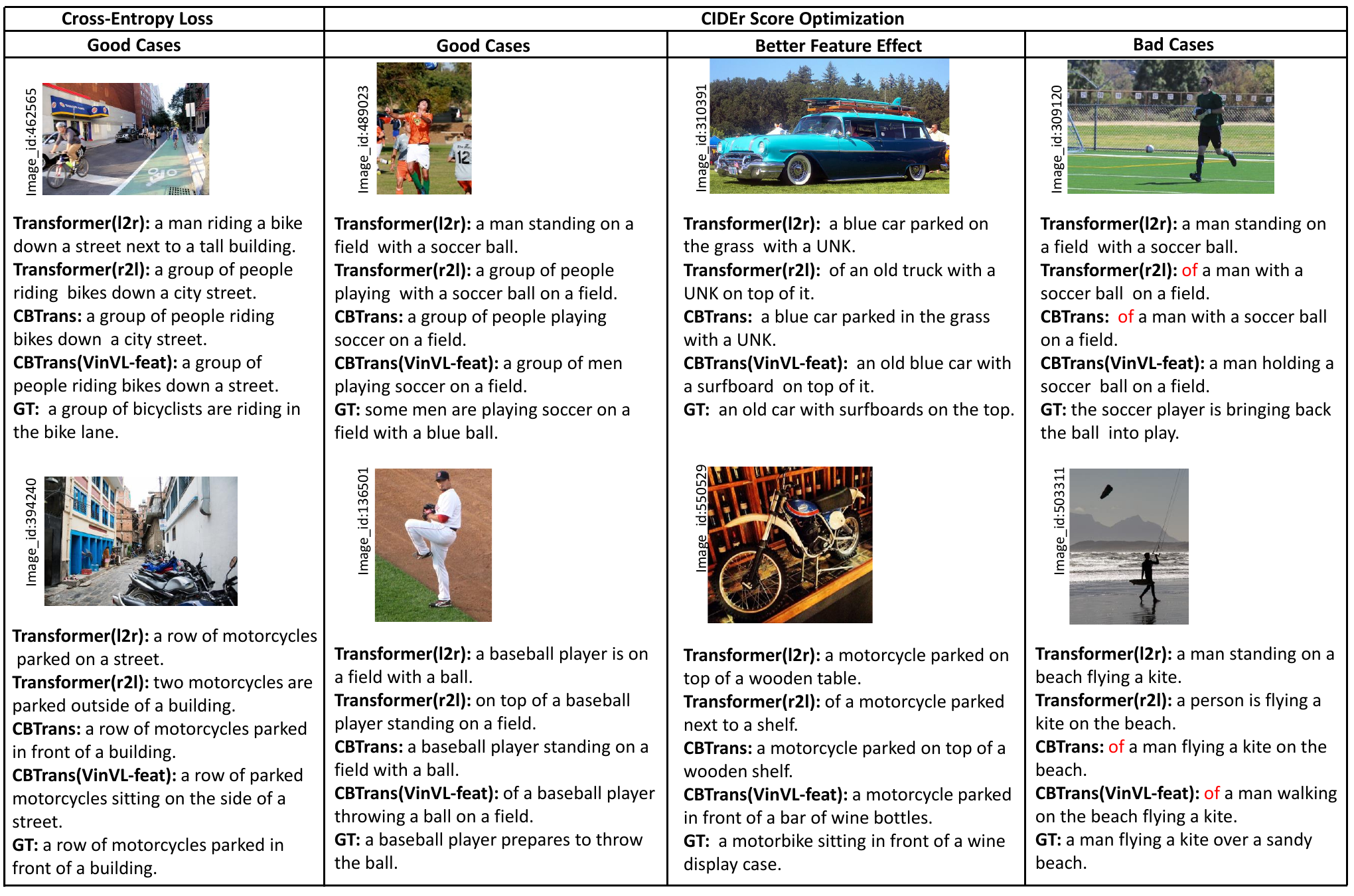} 
    \vspace{-0.3cm}	
	\caption{Examples of captions generated by our CBTrans model, conventional unidirectional Transformer model and human-annotated ground truth. Some bad  words are marked in red.}
	\label{fig:sample-CBTrans}
	\vspace{-0.6cm}	
\end{figure*}

\begin{figure}[ht]
	\centering
	\includegraphics[width=0.48\textwidth]{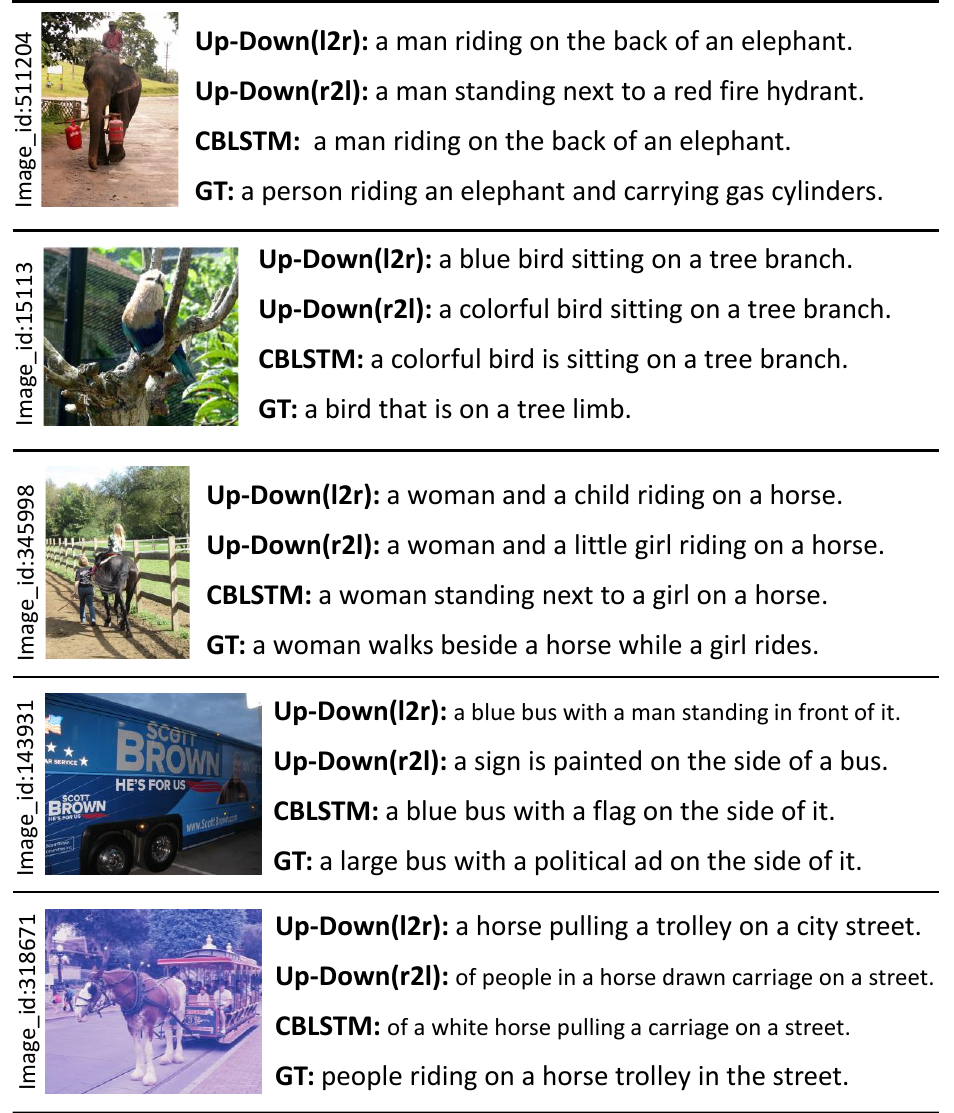} 
    \vspace{-0.3cm}	
	\caption{Examples of captions generated by our CBLSTM model, conventional unidirectional Up-Down model and human-annotated ground truth.}
	\label{fig:sample-CBLSTM}
	\vspace{-0.6cm}	
\end{figure}
Additionally, we also experimentally construct other two sentence-level ensemble models during evaluation to check the generality of this mechanism, where one is an ensemble of two Transformer(L2R) models and the other one is an ensemble of Transformer~(L2R) and Transformer~(R2L) models.
By observing the 3rd/6th/12th/15th rows of Table~\ref{tab:ablation-study-CBTrans}, we can find that this mechanism can still bring non-trivial improvement on METEOR metric but at the cost of saving and running two models.

Finally, we also report the performance comparison between our compact bidirectional model and the unidirectional counterpart  after CIDEr score optimization.  
Basically, CBTrans model has a similar advantage as in the first training stage as shown in Table~\ref{tab:comparison-CBTrans}, e.g., improving the conventional Transformer(L2R) model from 127.6~(22.6) to 129.8~(22.8) in CIDEr~(SPICE) metric. And the advantage is more obvious when using better VinVL feature by improving CIDEr~(SPICE) metric from $134.9$~($23.6$) to  $137.5$~($24.2$). 

CBLSTM model also holds this advantage as shown in Table~\ref{tab:comparison-CBLSTM}, e.g., improving the conventional UpDown(L2R) model from 130.6~(22.5) to 133.5~(22.9) in CIDEr~(SPICE) metric. However, we find that R2L 'branch' shows an undesirable phenomenon after CIDEr score optimization, i.e., generating some bad endings, e.g.'\textit{of a man with a soccer ball on a field}'. 
This undermines the overall model to some extent and causes degradation in BLEU metrics~(e.g., from 39.0 to 38.4 in Table~\ref{tab:comparison-CBLSTM}), which focus on n-gram matching.
Fortunately, this behavior can be eliminated by using removing bad endings trick, as shown in the bottom part of Table~\ref{tab:comparison-CBLSTM}.

\subsubsection{How does our models perform compared with state-of-the-art models?}
We show the performance comparisons between our models and  state-of-the-art models on 'Karpathy' test split in Table~\ref{tab:offline}. The performances of the single model and model ensemble are separately reported. The training manners of the Cross-Entropy Loss and CIDEr Score Optimization are also separately reported. The implementation of model ensemble follows the common practice~\cite{rennie2017self,huang2019attention}, which averaging the word-level output probability distributions of four independently trained instances with different parameter initialization.
In general, our CBTrans model exhibits better performance than other models except for X-LAN~\cite{pan2020x} and  RSTNet~\cite{zhang2021rstnet} in BLEU metrics in the single model setting.  In the model ensemble setting, our CBTrans model outperforms all other models in all metrics, especially a large margin in CIDEr~(about $5$\%). We think this is partly due to our CBTrans model can simultaneously take advantage of both word-level ensemble and sentence-level  ensemble in ensemble setting. In addition, we also report the performance of our ensemble model on the online testing server. Table~\ref{tab:online} details the performance over official testing  images with $5$ reference captions~(c5) and $40$ reference captions~(c40). The results clearly show that our CBTrans model shows better performance across all metrics, e.g., making the absolute improvement over the best competitor RSTNet by 4.1\%/4.6\% in CIDEr c5/c40.
It is noteworthy that we don't list recent vision-language pre-training models~\cite{zhou2020unified,li2020oscar,zhang2021vinvl} for comparison since it's not fair to directly compare them with non-pretraining-finetuning models.

\subsection{Qualitative Results}

We showcase some qualitative results generated by our CBTrans model and conventional unidirectional Transformer model, coupled with human-annotated ground truth sentences~(GT) in Fig.~\ref{fig:sample-CBTrans}. On average, CBTrans model can steal from both Transformer(l2r) and Transformer(r2l) models. For example, the CBTrans model absorbs the '\textit{a row of}' of Transformer~(l2r) and the '\textit{of a building}' of Transformer~(r2l) and  generates a caption that is closer to ground truth in the bottom-left example. In the top-left example, CBTrans  seems to directly choose the output of Transformer~(r2l), which is the better one. We  illustrate the effect of using a better feature in the 3rd column, e.g., CBTrans model fed with VinVL feature recognizes '\textit{surfboard}' and '\textit{a bar of wine bottles}'.
In the 4th column, we also show a kind of representative bad case of CBTrans model after CIDEr score optimization. This bad endings issue mainly derives from Transformer~(r2l) 'flow'. This is probably due to some prepositions~(e.g.,'\textit{of}') frequently follow '\textit{a}' in the reverse version of ground truth captions. This issue may be alleviated by using trick of removing bad endings or adding BLEU metric to the score function during self-critical training. We also showcase some qualitative results of CBLSTM model in Fig.~\ref{fig:sample-CBLSTM}. The first three are good cases and the last two are bad cases. Basically, CBLSTM model has similar advantage as CBTrans model.

\section{Conclusion}
\label{Conclusion}
In this paper, we introduce a compact bidirectional Transformer model for image captioning~(CBTrans) that can leverage bidirectional context implicitly and explicitly and the decoder can be executed parallelly. We conduct extensive ablation studies on MSCOCO benchmark and find that the compact architecture, which serves as a regularization for implicitly exploiting bidirectional context, and the sentence-level ensemble play more important roles than the explicit interaction mechanism. We further propose to combine word-level and sentence-level ensemble seamlessly and extend the conventional  self-critical training  to achieve new state-of-the-art results in comparison with non-vision-language-pretraining models. Finally, we verify the generality of this compact bidirectional architecture by extending it to LSTM backbone. 
The proposed compact bidirectional architecture is orthogonal to vision-language pretraining approaches, which focus on large-scale cross-modal pretraining. Our decoder could be integrated into vision-language pretraining frameworks by replacing their unidirectional decoders with our bidirectional design for further studies, potentially enhancing caption quality through better context utilization while retaining the benefits of pretrained representations.


\bibliographystyle{IEEEtran}
\bibliography{cbt}

\begin{IEEEbiography}[{\includegraphics[width=1in,height=1.25in,clip,keepaspectratio]{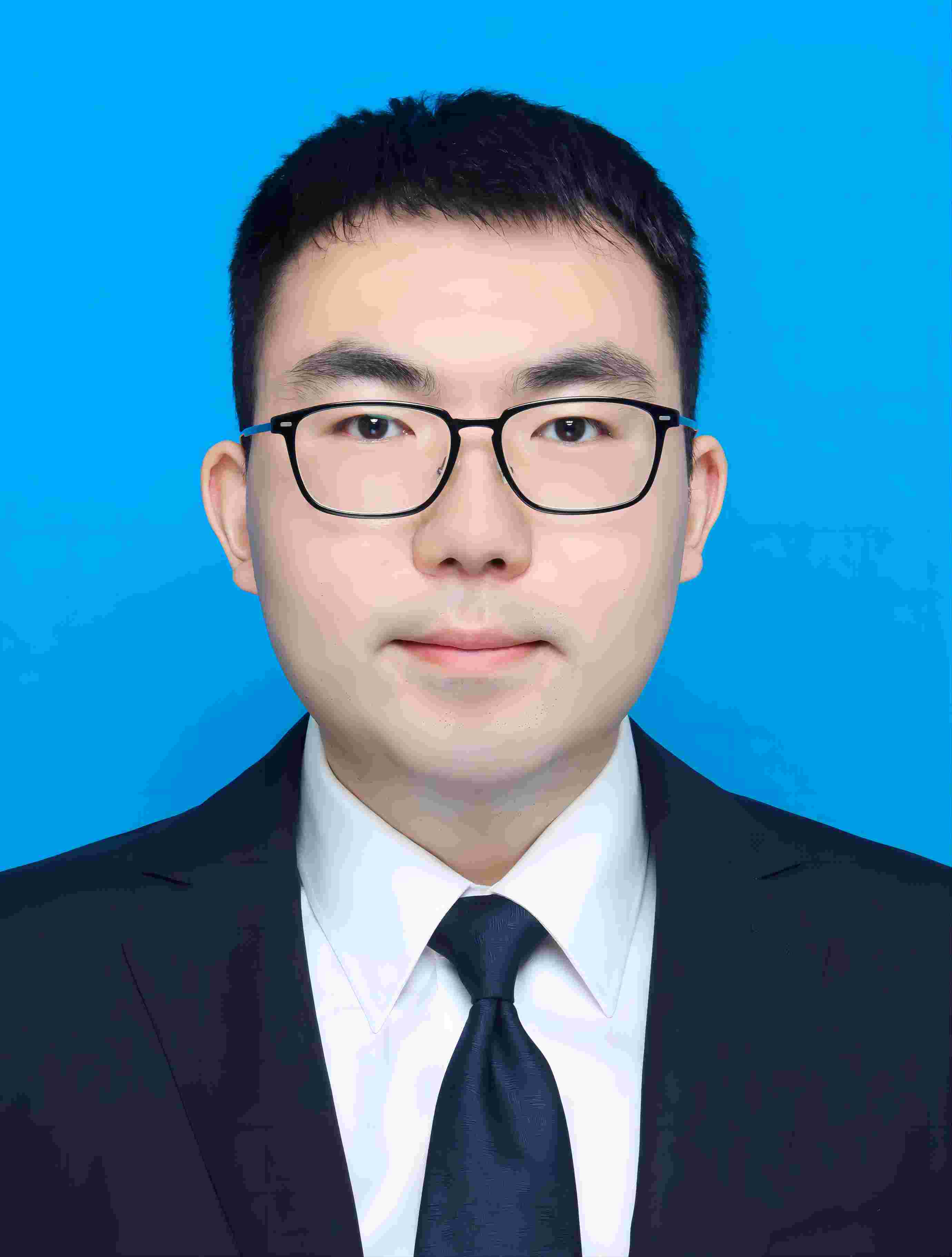}}]{Zijie Song} received the B.S. degree from North China Electric Power University, Beijing, China, in 2019. He received the M.E. degree and Ph.D. degree from Hefei University of Technology, Hefei, China, in 2022 and 2025. He currently serves as a lecturer with the School of Big Data and Statistics, Anhui University, Hefei, China. His research interests include cross-media analysis and reasoning.
\end{IEEEbiography}

\begin{IEEEbiography}[{\includegraphics[width=1in,height=1.25in,clip,keepaspectratio]{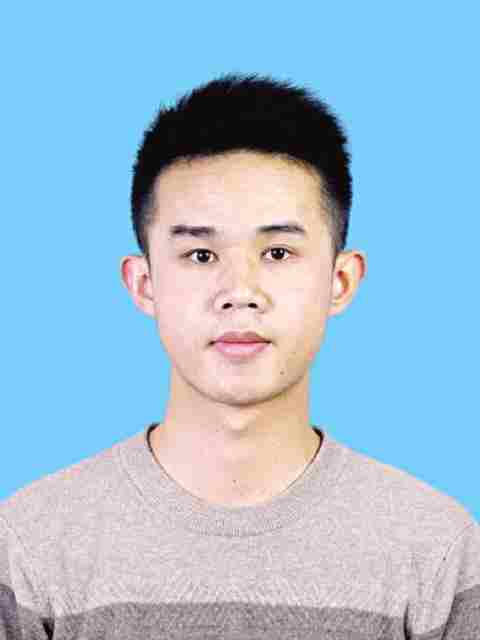}}]{Yuanen Zhou} received the Ph.D. degree from Hefei University of Technology (HFUT), Hefei, China, in 2022. He is currently a Special Associate Research Fellow of the Hefei Comprehensive National Science Center, Institute of Artificial Intelligence. His research interests include Vision\&Language, physiological signal processing, and affective computing.
\end{IEEEbiography}

\begin{IEEEbiography}[{\includegraphics[width=1in,height=1.25in,clip,keepaspectratio]{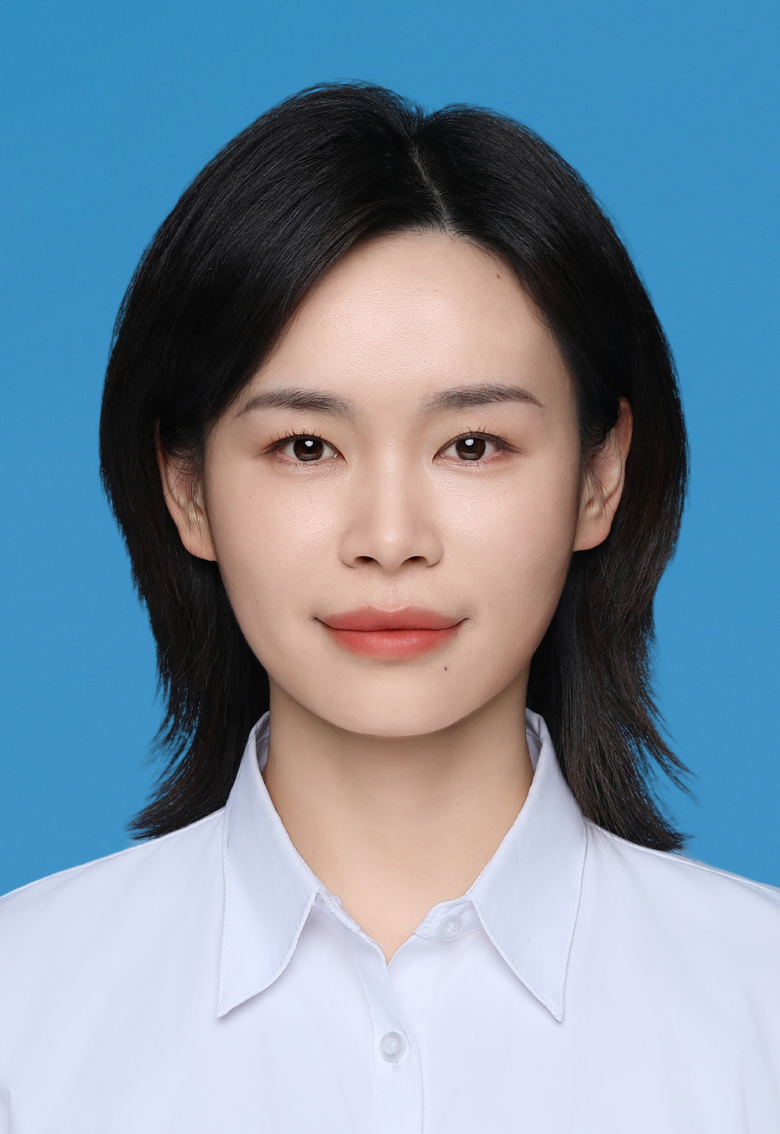}}]{Zhenzhen Hu}  received her Ph.D. degree from Hefei University of Technology, Hefei, China, in 2014. She is currently a full professor and Ph.D. Supervisor at the School of Computer Science and Information, Hefei University of Technology. Previously, she was a visiting student at the National University of Singapore and a Research Fellow at Nanyang Technological University, Singapore. Dr. Hu’s primary research interests include multimedia computing, computer vision, and image processing. She has authored more than 50 papers in leading international journals and conferences. In addition to her research contributions, Dr. Hu serves on the Women in Science Committee of the China Society of Image and Graphics, as well as the Multimedia Technical Committee and the Affective Computing Technical Committee.
\end{IEEEbiography}

\begin{IEEEbiography}[{\includegraphics[width=1in,height=1.25in,clip,keepaspectratio]{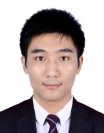}}]
{Daqing Liu} is an algorithm scientist with JD Explore Academy of JD.com Inc, China.
He received his PhD degree from the Department of Automation, University of Science and Technology of China (USTC) in 2021, under the supervision of Prof. Zheng-Jun Zha.
He used to be a research assistant with Nanyang Technological University (NTU) from 2018 to 2019, working with Assistant Prof. Hanwang Zhang.
His current research interest is multi-modality learning, including multi-modality understanding, retrieval, and generation.
\end{IEEEbiography}

\begin{IEEEbiography}[{\includegraphics[width=1in,height=1.25in,clip,keepaspectratio]{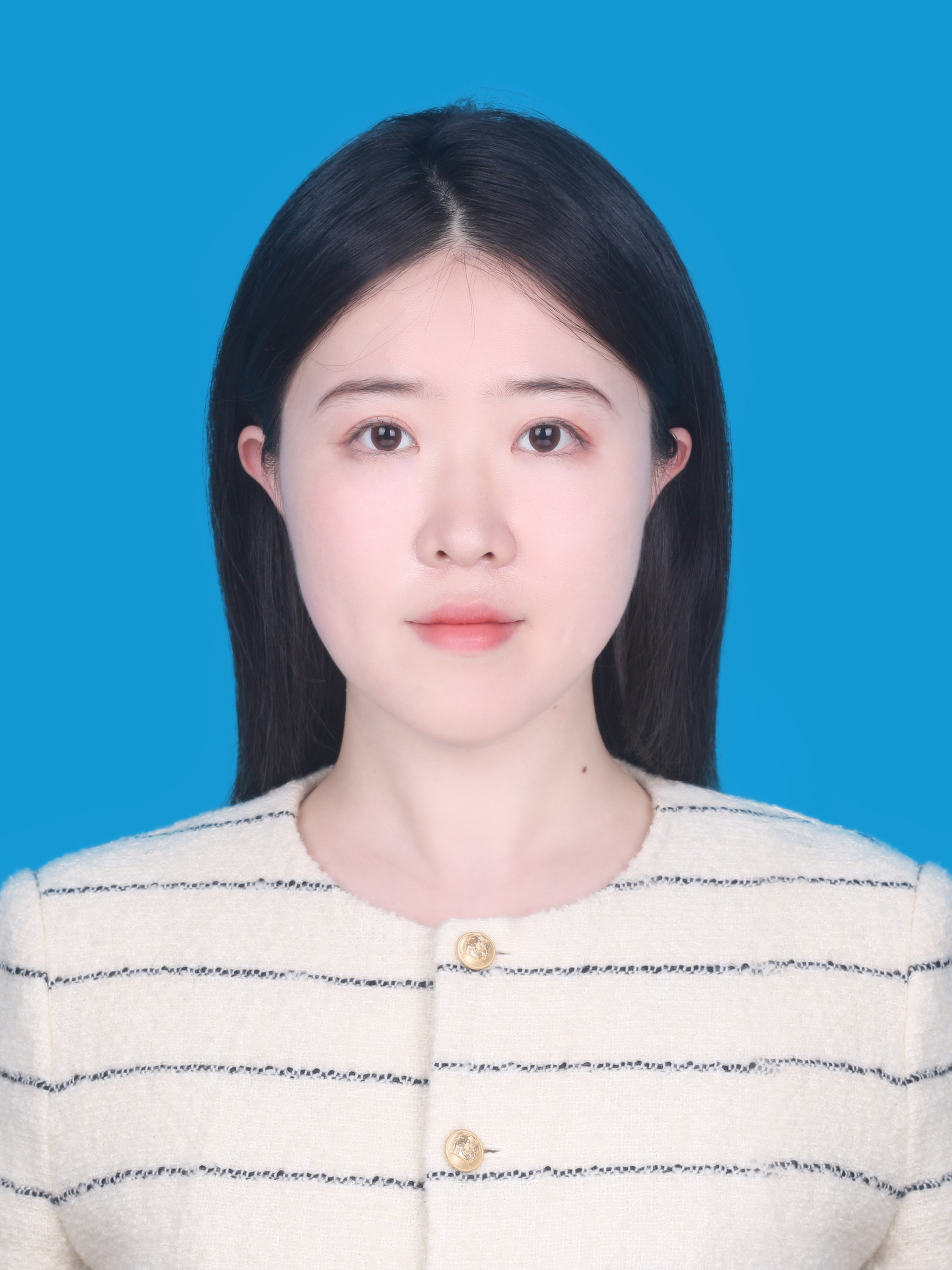}}]{Huixia Ben} received the B.E. and Ph.D. degrees from the Hefei University of Technology(HFUT), Anhui, China. She is currently a lecturer at State Key Laboratory of Digital Intelligent Technology for Unmanned Coal Mining, Anhui University of Science and Technology(AUST),China. Her research interests mainly include machine learning and multimediadata analysis, such as large-scale multimedia indexingand retrieval, multimedia data embedding, and visual understanding.
\end{IEEEbiography}

\begin{IEEEbiography}[{\includegraphics[width=1in,height=1.25in,clip,keepaspectratio]{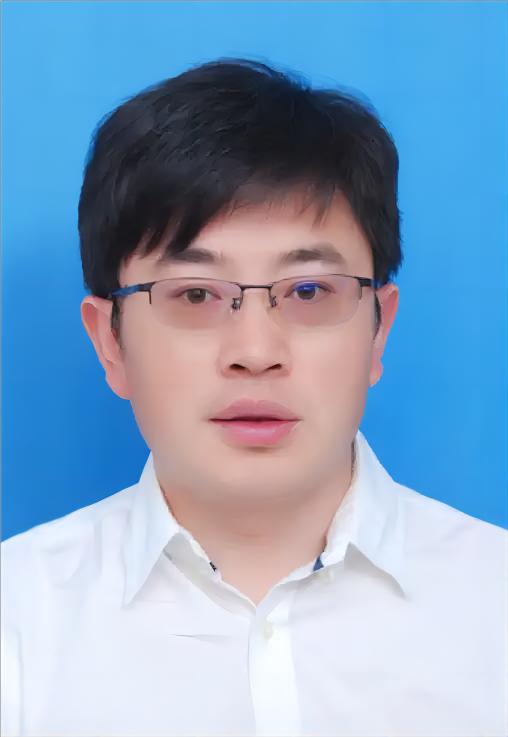}}]{Richang Hong} received the Ph.D. degree from the University of Science and Technology of China, Hefei, China, in 2008. He is currently a Professor with the Hefei University of Technology, Hefei. His research interests include multimedia content analysis and social media, in which he has coauthored more than 100 publications. He is a member of the ACM and an Executive Committee Member of the ACM SIGMM China Chapter. He was the Technical Program Chair of the MMM 2016, ICIMCS 2017, and PCM 2018. He was a recipient of the Best Paper Award at the ACM Multimedia 2010, the Best Paper Award at the ACM ICMR 2015, and the Honorable Mention of IEEE Transactions on Multimedia Best Paper Award 2015. He was an Associate Editor of IEEE Multimedia Magazine and Information Sciences and Signal Processing, Elsevier.
\end{IEEEbiography}

\begin{IEEEbiography}[{\includegraphics[width=1in,height=1.25in,clip,keepaspectratio]{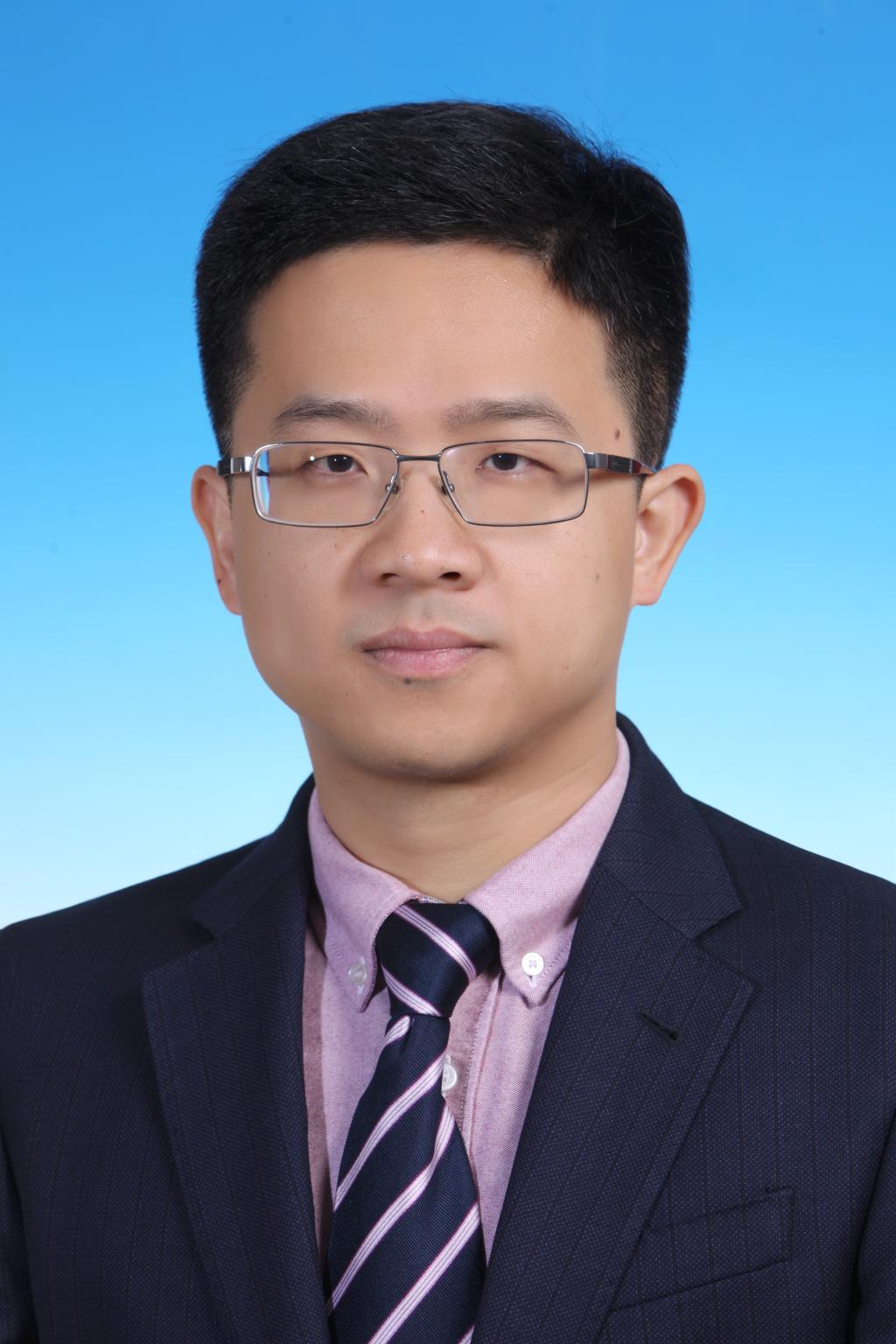}}]{Meng Wang} received the B.E. and Ph.D. degrees in signal and information processing with special class for the Gifted Young from the Department of Electronic Engineering and Information Science, University of Science and Technology of China, Hefei, China, in 2003 and 2008, respectively. He is currently a Professor with Hefei University of Technology, Hefei, China. He has authored over 200 book chapters, journals, and conference papers in his research areas. His current research interests include multimedia content analysis, computer vision, and pattern recognition. Dr. Wang was a recipient of the ACM SIGMM Rising Star Award 2014. He is an Associate Editor of IEEE Transactions on Knowledge and Data Engineering, IEEE Transactions on Circuits and Systems for Video Technology, IEEE Transactions on Multimedia, and IEEE Transactions on Neural Networks and Learning Systems.
\end{IEEEbiography}


\end{document}